\documentclass[11pt]{article}
\usepackage{amssymb}
\usepackage{amsmath}
\usepackage{amsthm}
\usepackage{graphics}
\usepackage{epsfig,amssymb,latexsym,verbatim}
\usepackage{graphicx}
\usepackage{multirow}
\usepackage{multicol}
\usepackage{float}
\usepackage{booktabs}
\usepackage{fancyhdr}
\usepackage{enumerate,verbatim}
\usepackage{rotating}
\usepackage{graphicx}
\usepackage{subfigure}
\usepackage{graphicx}
\usepackage{ulem}
\usepackage{url}
\usepackage{threeparttable}
\usepackage{multirow}
\usepackage{multicol}
\usepackage{wrapfig}
\usepackage{lscape}
\usepackage{rotating}
\usepackage{epstopdf}
\usepackage{subfigure}
\usepackage[plain,noend]{algorithm2e}
\usepackage[noblocks]{authblk}

\def\bs{\boldsymbol}

\newcommand{\ceil}[1]{\lceil{#1}\rceil}
\newtheorem{theorem}{{\sc Theorem}}

\begin{document}
%

\title{Estimation of the Mean Function of Functional Data via Deep Neural Networks}


\author[1]{Shuoyang Wang}
\author[1]{Guanqun Cao}
\author[2]{Zuofeng Shang}

\affil[1]{Department Mathematics and Statistics, Auburn University, U.S.A.}
\affil[2]{Department of Mathematical Sciences, New Jersey Institute of Technology, U.S.A.}

\date{}

\maketitle

\begin{quotation}
\noindent \textit{Abstract:} In this work, we propose a deep neural network method to perform nonparametric regression for
  functional data.  The proposed estimators are based on sparsely connected deep neural networks with ReLU activation function. By properly choosing network
architecture, our estimator achieves the optimal nonparametric convergence rate in empirical norm. Under certain circumstances such as
trigonometric polynomial kernel and a sufficiently large sampling frequency, the convergence rate
is even faster than root-$n$ rate. Through
Monte Carlo simulation studies we examine the finite-sample performance of the proposed method. Finally, the
proposed method is applied to analyze positron emission tomography images of patients with Alzheimer disease obtained from the Alzheimer Disease Neuroimaging Initiative database.

\vspace{9pt} \noindent \textit{Key words and phrases:} Functional data analysis; multilayer sparse neural networks; nonparametric regression; rate of convergence; ReLU activation function.
\end{quotation}

%

\pagestyle{myheadings}
\thispagestyle{plain}
{}


\section{Introduction}
Functional data refer to curves or functions, i.e. the data for each variable are
viewed as smooth curves, surfaces, or hypersurfaces evaluated at a finite subset
of some interval in  1D  and 2D  (e.g., some period of time, some range of pixels or voxels   and so on). Functional data means intrinsically infinite-dimensional
but are usually measured discretely.  The high intrinsic dimensionality of these data poses challenges both for
theory and computation. Functional data analysis (FDA) has been a topic of increasing interest in the statistics community for recent decades. \cite{Ramsay:Silverman:05} and \cite{Wang:etal:16} gave a comprehensive overview of FDA. The atom of functional data is a function,
where for each subject in a random sample, one or several functions are recorded.
It consists of a collection of independent and identical realizations $\left\{
\xi _{i}(x)\right\} _{i=1}^{n}$ of a smooth random function $\xi (x)$,
with unknown mean function $E\xi (x)=f(x)$ and covariance function $G\left(
x,x^{\prime }\right) =\mbox{cov}\left\{ \xi (x),\xi (x^{\prime })\right\} $%
. Although the domain of $\xi (\cdot )$ is an entire interval $\mathcal{X}$%
, the recording of each random curve $\xi_{i}\left( x\right) $ is only
over a finite number $N_{i}$ of points in $\mathcal{X} \in \mathbb{R}^d$, $d=1,2,\ldots$, and contaminated
with measurement errors.

\subsection{Related literature}
In FDA problems, estimation of mean functions $f(x)$ is the
fundamental first step; see  \cite{cardot:00,Rice:Wu:01,ferraty2006nonparametric} for example. Various methods exist that allow to estimate the regression function nonparametrically. \cite{Rice:Wu:01}  adopted the mixed effect models where the mean function and the eigenfunctions were represented with B-splines and the spline coefficients were estimated by the EM algorithm; \cite{Yao:etal:05b} applied the local linear smoothers   to estimate the mean and the covariance functions.   \cite{Morris:Carroll:06} generalized the linear mixed model to the functional mixed model framework, with model fitting done by using a Bayesian wavelet-based approach. In \cite{Cao:Yang:Todem:12}, a polynomial spline estimator is proposed for the mean function of functional data together with a simultaneous confidence band. These nonparametric methods apply the pre-specified basis expansion, e.g., polynomial spline, local linear smoother, wavelet and so on, to fit the
unknown mean function. The convergence rates achieve either optimal nonparametric rate or parametric rate dependents on how dense of the observed points for each subject.

Even though FDA has received considerable attention over the last decade, most approaches still focus on 1D functional data.  The high intrinsic dimensionality of these data poses challenges both for theory and computation; these challenges vary with how the functional data were sampled. Hence,  few are developed for general high-dimensional functional data. Recently,   several attempts have been made to extend these nonparametric methods for spatial and image data.  \cite{Wang:Wang:Wang:Ogden:19}  used bivariate splines over triangulations to handle an irregular domain of the images that is common in brain imaging studies.  The proposed spline estimators of the mean functions are shown to be consistent and asymptotically normal.  However, the triangularized bivariate splines are designed for 2D functions only. Extending spline basis functions for general $d$-dimensional data observed on an irregular domain is very sophisticated and becomes extremely complex as $d$ increases.  \cite{Wang:Nan:Zhu:Koeppe:14}  proposed a regularized Haar wavelet-based approach for the analysis of 3D  brain image data in the framework of functional linear regression model.

 Another popular method is functional principal component analysis (FPCA)  which is an extension of multivariate
principal component analysis, see \cite{Hall:etal:06, Yao:etal:05a}  for example. Recently, there are a few studies on 2D FDA. \cite{Zhou:Pan:14} proposed a smooth FPCA for
2D functions on irregular planar domains; their approach is based on
a mixed effects model that specifies the principal component functions as bivariate splines on triangulations and the principal component scores as random effects. \cite{Lila:etal:16} proposed a FPCA model
that can handle real functions observable on a 2D manifold.  \cite{Chen:Jiang:17} extended it to analyze functional/longitudinal data observed on a general $d$-dimensional
domain. They showed that the proposed estimators
can achieve the classical nonparametric rates for longitudinal data and the optimal convergence rates for functional
data if the number of observations per sample is of the order $(n / \log n)^{d/4}$. There are several issues when applying FPCA. One is to choose the form of
the orthonormal eigenfunctions. Note that any functions can be represented by its
orthogonal bases. The choice of the basis decides the shape of the curve. Another issue is to choose the number of eigenfunctions.
This is an important practical issue without a satisfactory theoretical solution. Presumably, the
larger  the number of eigenfunctions, the more flexible the approximation would be, and hence, the closer
to the true curve. However, a large   number of eigenfunctions always result in a complex model which
introduces difficulties to follow-up analysis.

For many years, the use of neural networks has been one of the most promising approaches in connection with applications related to approximation and estimation of multivariate functions  (see, e.g., \cite{Anthony:Bartlett:09,Ripley:14}). Recently, the focus is on multilayer neural networks, which use many hidden layers, and the corresponding techniques are called {\it deep learning}.  Under  the nonparametric regression model,  via sparsely connected deep neural networks, \cite{Schmidt:19} showed that the $L_2$ errors of the least squares neural network regression estimator  achieves the same minimax   rate of convergence (up to a logarithmic factor) as proposed in \cite{Stone:82}. Furthermore, this neural network estimator does not suffer  the curse-of-dimensionality which is a classical drawback in the traditional nonparametric regression framework.  \cite{Bauer:Kohler:19} has also obtained the similar results under deep learning frame work via a different activation function.
\cite{Liu:19} further removed the logarithmic factors to achieve exact optimal nonparametric rate.

\subsection{Our contributions}

Our  major contribution is resolving the curse-of-dimensionality and model misspecification issues in high-dimensional FDA by borrowing the advantage from the  {\it deep learning} domain.
To our best knowledge, most existing methods for estimating the mean function  in high-dimensional FDA suffers at least one of the two major issues. The first issue is the curse-of-dimensionality. When the observed
points come from a hypercube, i.e., $[0,1]^d$, $d=3$ for 3D imaging study, the nonparametric convergence rates are slower than the optimal nonparametric rate. This means that no statistical procedure can perfectly recover the signal pointwisely. The second concern is the misspecification of the true model and complexity of the imposed model.  Since the only method to circumvent the curse-of-dimensionality is to assume additional structure assumptions, for example, additive models and single-index models, on the target function to achieve better rates of convergence. These structured models can derive optimal convergence rates only if the imposed structure are satisfied. Therefore, it is useful to derive rates of convergence given more general types of functions, which is highly demanding in real applications.
As suggested by \cite{Bauer:Kohler:19}, ``{\it the curse-of-dimensionality issue can be resolved when the true regression functions  are constructed in a modular form,  where each modular part computes a function depending only on a few of the components of the high-dimensional input.  At the meanwhile the modularity of the system can be extremely complex and deep, which resolves the misidentification issue.}'' Motivated by  these attracting features of the deep neural network, we conduct  FDA in the {\it deep learning} domain to overcome the curse-of-dimensionality and model misspecification issues.

Denote by $Y_{ij}$ the $j$-th observation of the random curve $\xi
_{i}(\cdot )$ at grid points $\mathbf{X}_{ij}$, $1\leq i\leq n,1\leq j\leq N_{i}$.
  In
this paper, for simple notations, we examine the equally spaced   design, in other words, $%
\mathbf{X}_{ij}=\mathbf{X}_j=j/N,1\leq i\leq n,1\leq j\leq N$ with $N$ going to infinity. The main results can be extended to irregularly spaced design.  For the $%
i$-th\ subject, $i=1,2,...,n$, its sample path $\left\{\mathbf{X}_j,Y_{ij}\right\} $
consists of the noisy realization of the Gaussian process $\xi
_{i}(\mathbf{X})$\ in the sense that $Y_{ij}=\xi _{i}\left( \mathbf{X}_j\right) + \epsilon _{i}( \mathbf{X}_j) $, and $\left\{ \xi _{i}(\mathbf{X}), \mathbf{X} \in \left[ 0,1\right]^d \right\} $
are i.i.d. copies of the process $\left\{ \xi (\mathbf{X}), \mathbf{X}\in \left[ 0,1\right]^d
\right\} $ which is $L^{2}$, i.e., $E\int_{\left[ 0,1\right] ^d }\xi
^{2}(\mathbf{X})d\mathbf{X}<+\infty $.  The error term $\epsilon _{i}( \mathbf{X}_j)$
has   mean zero and finite variance,
In this work, we consider fitting a feedforward neural network
to the functional data. Under standard conditions in FDA literature, the proposed neural network estimator has convergence rate (in empirical norm)
\begin{equation}\label{main:rate:conv}
(nN^{\varrho})^{-\frac{2\beta^{\ast}}{2\beta^{\ast}+t^{\ast}}}\log^6 (nN^{\varrho}),
\end{equation}
where $\beta^{\ast}>0$ characterizes the smoothness of the modular components of the true function, $t^{\ast}\in(0,d)$
is the intrinsic dimension of the true function, and $\varrho\ge0$ is the decay rate of the maximal eigenvalue of the covariance matrix.
An interesting finding is that, with $\varrho>0$, (\ref{main:rate:conv}) can be even
faster than $n^{-1}$ when $N\gg n^{\frac{t^*}{2\beta^*\varrho}}$. In other words,
our neural network estimator is ``superconvergent'' similar as the smoothing spline estimator considered by \cite{CY:11}.

Different from the existing neural network literature on nonparametric regression \cite{Bauer:Kohler:19,Schmidt:19}, which only handle i.i.d. data, we focus on FDA, where each subject is an  random curve in a hypercube. Because of this special data structure, the major challenge  becomes to deal with  the correlation among the $N$ evaluation points in the framework of neural network, which  has been done in the existing works. It is not surprising that the convergence rate increase with $n$ (the number of independent realizations), since the realizations are i.i.d., but we also derive the convergence rate also increases with $N$. Furthermore, under some realistic conditions, the rate of convergence can be even faster than the optimal root-$n$ rate, which has not been discussed clearly in any FDA literature yet.

The paper is structured as follows. Section \ref{SEC:Math} introduces multilayer feedforward
artificial neural networks and discusses mathematical modeling.
This section also contains the definition of the network classes. Section \ref{SEC:FDA} provides the model setting in FDA. The considered
function classes for the regression function and the main result can
be found in Section \ref{SEC:main}.  In Section \ref{SEC:sim}, it is shown that the finite sample performance of  proposed neural network estimator. The proposed method is applied to the spatially normalized positron emission tomography (PET) data from Alzheimer Disease Neuroimaging Initiative (ADNI) in Section \ref{SEC:realdata} and make some concluding remarks in Section \ref{SEC: discussion}. The proof of the
main result together with additional discussion can be found in the Supplementary material.

\section{Review of ReLU Feedforward Neural Network}
\label{SEC:Math}

In the feedforward neural network,  the  activation function $\sigma$ and the network architecture are two important components that impact the asymptotic and non-asymptotic properties of the target functions. Motivated by the importance in deep learning and its recent applications in statistical nonparametric regression modeling \cite{Schmidt:19}, we study the rectifier linear unit (ReLU) activation function
\[
\sigma(x)=\max(x,0).
\]
For any vector $\mathbf{v}=(v_1,\ldots,v_d)\in \mathbb{R}^d$, define the shifted activation function $\sigma_{\mathbf{v}}: \mathbb{R}^d \rightarrow \mathbb{R}^d$ as:
\[
\sigma_{\mathbf{v}}\left(\begin{array}{c} y_1\\ \vdots \\y_d \end{array} \right)  = \left(\begin{array}{c} \sigma(y_1-v_1)\\ \vdots \\ \sigma(y_d-v_d) \end{array} \right).
\]
The network architecture $(L, \mathbf{p})$ consists of a positive integer $L$ called the number of hidden layers and a width vector $\mathbf{p} = (p_0, \ldots, p_{L+1}) \in \mathbb{N}^{L+2}$. A feedforward neural network with network architecture $(L, \mathbf{p})$ is then any function of the form
\begin{equation} \label{EQ:f}
f: \mathbb{R}^{p_0} \rightarrow \mathbb{R}^{p_{L+1}},\;\;\;  \mathbf{x}\mapsto f(\mathbf{x}) = \mathbf{W}_l\sigma_{\bs{V}_L} W_{L-1}\sigma_{\bs{V}_{L-1}}\ldots \mathbf{W}_1\sigma_{\bs{V}_1} \mathbf{W}_0\mathbf{x},
\end{equation}
where $\mathbf{x} \in \mathbb{R}^d$, $\mathbf{W}_l$ is a $p_l \times p_{l+1}$ weight matrix and $\bs{v}_l \in \mathbb{R}^{p_l}$ is a shift vector. To fit networks with data generated from the $d$-dimensional hypercube functional data model, we must have $p_0 = d$ and $p_{L+1} = 1$. 

Given a network function (\ref{EQ:f}),
 the entries of the matrices $\mathbf{W}_l$ and vectors $\mathbf{v}_l$, $l=1,\ldots,L$, are the unknown network  parameters. These parameters need to be estimated from the data.
Define $ \| \mathbf{W}_l\|_{\infty}$ as the maximum-entry norm of $\mathbf{W}_l$. The space of network functions with given network architecture and network parameters bounded by one, i.e.,
\begin{eqnarray*}
  \mathcal{F}(L, \mathbf{p} )
= \left\{ f(\cdot) \mbox{  of~ the~ form~} (\ref{EQ:f}) : \max_{l=0,\ldots,L}\| \mathbf{W}_l\|_{\infty} + |\mathbf{v}_l|_{\infty} \leq 1 \right\},
\end{eqnarray*}
and for simplicity reasons, let $\mathbf{v}_0$ be a zero vector.

In deep learning, sparsity of the neural network is enforced through regularization or specific forms of networks. Dropout for instance sets randomly units to zero and has the effect that each unit will be active only for a small fraction of the data (Section 7.2 in \cite{Srivastava:etal:14}). In this work, we model the network sparsity assuming that there are only few non-zero/active network parameters. The $s$-sparse networks for our functional data model are given by
\begin{equation}\label{EQ:class}
\mathcal{F} = \mathcal{F}(L, \mathbf{p}, s, F) = \left\{ f \in \mathcal{F}(L, \mathbf{p}) : \sum_{l=0}^L\| \mathbf{W}_l\|_0 + |\mathbf{v}_l|_0 \leq s, \| f\|_N \leq F\right\},
\end{equation}
where $\| \mathbf{W}_l\|_0$ denotes the number of non-zero entries of $\mathbf{W}_l$ and  the empirical norm $\| f\|_N$ is defined by $\| f\|_N = \left(\frac{1}{N}\sum_{j=1}^Nf^2(\mathbf{x}_j)\right)^{1/2}$. Note that $\mathcal{F}(L, \mathbf{p})$
is broader than the one considered by \cite{Schmidt:19} who assume the supnorm of the network functions
to be bounded.

The theoretical performance of neural network highly depends on the underlying function class. Analogous to \cite{Schmidt:19}, we assume the true mean function $f_0$ is a composition of several functions:
\begin{equation*}
f_0 = g_q \circ g_{q-1} \circ \ldots \circ g_1 \circ g_0,
\end{equation*}
with $g_i : \left[a_i, b_i \right]^{d_i} \rightarrow \left[a_{i+1}, b_{i+1} \right]^{d_{i+1}} $, where $g_i = \left( g_{ij}\right)^{\top}_{j=1, \ldots, d_{i+1}}$, $i=1,\ldots,q$. Let $t_i$ be the maximal number of variables on which each of the $g_{ij}$ depends on, and $t_i$ might be much smaller than $d_i$. This function class is natural for neural networks.
Define the ball of $\beta$-H\"{o}lder functions with radius $K$ as
\begin{eqnarray*}
\mathcal{C}_d^{\beta}(D, K) = &&\left\{ \right.  f: D \subset \mathbb{R}^d \rightarrow \mathbb{R} : \\
&&\sum_{\bs{\alpha}:|\bs{\alpha}|<\beta} \| \partial^{\bs{\alpha}}f\|_{\infty} + \sum_{\bs{\alpha}:|\bs{\alpha}|=\lfloor{\beta}\rfloor}\sup_{\mathbf{x}, \mathbf{y} \in D, \mathbf{x} \neq \mathbf{y} } \frac{|\partial^{\bs{\alpha}}f(\mathbf{x}) - \partial^{\mathbf{\alpha}}f(\bs{y}) |}{|\mathbf{x} - \mathbf{y}|_{\infty}^{\beta - \lfloor\beta\rfloor}} \leq K \left.\right\},
\end{eqnarray*}
where $\partial^{\bs{\alpha}}$ = $\partial^{\alpha_1}\ldots\partial^{\alpha_d}$ with $\bs{\alpha}$ = $(\alpha_1, \ldots, \alpha_d) \in \mathbb{N}^d$ and $|\bs{\alpha}|:= |\bs{\alpha}|_1$.
We assume each  $g_{ij}$  is $\beta_i$-H\"{o}lder function with radius $K_i$. Since $g_{ij}$ is also $t_i$-variate,   the underlying function space becomes
\begin{eqnarray}\label{EQ:gfunction}
\mathcal{G}\left(q, \mathbf{d}, \mathbf{t}, \bs{\beta}, \mathbf{K} \right) := \left\{ \right. f = g_q \circ \ldots \circ g_0 : && g_i = (g_{ij})_j : \left[a_i, b_i \right]^{d_i} \rightarrow \left[a_{i+1}, b_{i+1} \right]^{d_{i+1}}, \nonumber \\
&& g_{ij} \in \mathcal{C}^{\beta_i}_{t_i}\left( \left[a_i, b_i\right]^{t_i}, K_i\right), |a_i|, |b_i| \leq K_i \left. \right\},
\end{eqnarray}
with $\mathbf{d} := (d_0, \ldots, d_{q+1})$,  $\mathbf{t} := (t_0, \ldots, t_q)$,  $\bs{\beta} := (\beta_0, \ldots, \beta_q)$, $\mathbf{K} := (K_0, \ldots, K_q)$ and
$
\beta_i^{\ast} := \beta_i \prod_{k=i+1}^q (\beta_k \wedge 1)
$.

\section{Functional Data Analysis Model  }
\label{SEC:FDA}

In this work, we consider the following classical FDA model:
\begin{eqnarray*}
Y_{ij} &=&\xi_i\left(\mathbf{X}_j\right) + \epsilon_{i}\left(\mathbf{X}_j\right)\\
  &= &f_0\left(\mathbf{X}_j\right) + \eta_i\left(\mathbf{X}_j\right) + \epsilon_{i}\left(\mathbf{X}_j\right), ~~i = 1, 2, \ldots, n, j = 1, 2, \ldots, N,
\end{eqnarray*}
where $f_0: \mathbb{R}^d\rightarrow \mathbb{R}$, $E(Y_{ij})=f_0\left(\mathbf{X}_j\right)$, $n$ is the sample size, $N$ is the total number of observation points in a $d$-dimensional hypercube, i.e., $\mathbf{X}_j=(X_{j1}, \ldots, X_{jd})$. Without loss of generality, let $\mathbf{X}_j \in\left[ 0,1\right]^d$, $j=1,\ldots,N$. Note that the main results can be easily extended to the scenario with irregularly spaced design, i.e., $\mathbf{X}_j$ replaced by  $\mathbf{X}_{ij}$, and $N$ replaced by $N_i$ for each $i$.
Let $\rho_{ij} =  \eta_i\left(\mathbf{X}_j\right) + \epsilon_{i}\left(\mathbf{X}_j\right)$, where $\eta_i$ is a Gaussian process characterizing individual curve variations from $f_0\left(\cdot\right)$, and it has mean zero and $\mathrm{Cov}(\eta(\mathbf{X}_j) ,\eta(\mathbf{X}_{j'})):=G( \mathbf{X}_j,  \mathbf{X}_{j'})$, and $ \epsilon_{i}\left(\mathbf{X}_j\right) =\tau \left(\mathbf{X}_j\right) \varepsilon _{ij}$, where $\varepsilon _{ij}$'s are independent normal random variables and $\tau(\mathbf{X})$ is the standard deviation function    bounded above zero for any $\mathbf{X} \in \left[ 0, 1\right]^d$.
By Mercer's Theorem, covariance function $ G(\mathbf{x}, \mathbf{x'})$ has the following spectrum decomposition
\begin{equation*}
G(\mathbf{x}, \mathbf{x'}) = \sum_{k=1}^{\infty}\lambda_k\psi_k(\mathbf{x})\psi_k(\mathbf{x'}),
\end{equation*}
where $\left\{ \lambda_k \right\}_{k=1}^{\infty}$ and $\left\{ \psi_k(\mathbf{x}) \right\}_{k=1}^{\infty}$ are the eigenvalues and eigenfunctions of $G(\mathbf{x}, \mathbf{x'})$, respectively,
and $\left\{ \psi_k(\mathbf{x}) \right\}_{k=1}^{\infty}$ are orthonormal bases in $L_2(\left[0, 1\right]^d)$.

In the functional data regression model,  the common objective of all estimation  methods is to find an optimal estimator by least-square loss function. In the neural network, this coincides with
finding  networks $f$ with smallest empirical risk $ \frac{1}{N}\sum_{j = 1}^N\left\{\overline{Y}_{\cdot j} - f\left(\mathbf{X}_j\right)\right\}^2$, where $\overline{Y}_{\cdot j}  = \frac{1}{n}\sum_{j=1}^nY_{ij}$. The proposed deep neural network (DNN) estimator  is
\begin{equation*}
\widehat{f} = \arg\min_{f\in \mathcal{F}}  \frac{1}{N}\sum_{j = 1}^N\left\{\overline{Y}_{\cdot j}- f\left(\mathbf{X}_j\right)\right\}^2.
\end{equation*}
Define $f^{\ast} = \arg\min_{f\in \mathcal{F}}\| f_0 - f\|_{\infty}$. Note that
\begin{equation*}
\frac{1}{N}\sum_{j=1}^N\left( \overline{Y}_{\cdot j} - \widehat{f}(\mathbf{X}_j)\right)^2 \leq \frac{1}{N}\sum_{j=1}^N\left( \overline{Y}_{\cdot j}- f^{\ast}(\mathbf{X}_j)\right)^2,
\end{equation*}
which is equivalent to
$$
\frac{1}{N}\sum_{j=1}^N\left( f_0(\mathbf{X}_j) - \widehat{f}(\mathbf{X}_j)+\overline{\rho}_{\cdot j}\right)^2 \leq \frac{1}{N}\sum_{j=1}^N\left( f_0(\mathbf{X}_j)- f^{\ast}(\mathbf{X}_j)+\overline{\rho}_{\cdot j}\right)^2,$$where $\overline{\rho}_{\cdot j} = \frac{1}{n}\sum_{i = 1}^n\rho_{ij} = \frac{1}{n}\sum_{i = 1}^n  {\eta}_{i}\left(\mathbf{X}_j\right) + \frac{1}{n}\sum_{i = 1}^n {\epsilon}_{i} \left(\mathbf{X}_j\right)$.
Therefore, we have
\begin{equation}\label{EQ:Two parts}
 \frac{1}{N}\sum_{j=1}^N\left( \widehat{f}(\mathbf{X}_j) - f_0(\mathbf{X}_j)\right)^2 \leq    \frac{1}{N}\sum_{j=1}^N\left( f^{\ast}(\mathbf{X}_j) - f_0(\mathbf{X}_j)\right)^2 + \frac{2}{N}\sum_{j=1}^N\left(  \widehat{f}(\mathbf{X}_j)- f^{\ast}(\mathbf{X}_j)\right)\overline{\rho}_{\cdot j}.
\end{equation}
The above equation indicates that the empirical norm of the estimator
is bounded by two items. The first item is essentially determined by the distance between the network class $\mathcal{F}$ and true
 function class $f_0$ which can be arbitrarily small due to a result by \cite{Schmidt:19}. The second item is a weighted average of a random process, and it is affected by the parameters in both $\mathcal{F}$ and $\mathcal{G}(q, \mathbf{d},\mathbf{t}, \bs{\beta}, \mathbf{K})$, and the characteristic of the error terms.

\section{Main results: Convergence Rate of the Deep Neural Network Estimator}
\label{SEC:main}

For simple notations, $\log$ means the
 logarithmic function with base $2$.
For  sequences $(a_n)_n$ and $(b_n)_n$, we write $a_n \lesssim b_n$
if there exists a constant $C$ such that $a_n \leq Cb_n$ for all $n$. $a_n \asymp b_n$ means $a_n \lesssim b_n$  and $a_n \gtrsim b_n$.
Write $\lfloor x \rfloor$  for the largest integer $\leq x$ and $\lceil x \rceil$  for the smallest integer $\geq x$.
Let $\mathbf{C}_{N}=[G(\mathbf{X}_j,\mathbf{X}_{j'})/N]_{j,j'=1}^N$ as the $N$ by $N$ kernel matrix corresponding to covariance function $G$,
We now introduce the main assumptions of this article:
\begin{description}
\item[(A1)]  The true regression function $f_0 \in \mathcal{G}\left(q, \mathbf{d}, \mathbf{t}, \bs{\beta}, \mathbf{K} \right)$.
\item[(A2)]  The standard deviation function $\tau(\cdot)$ is bounded for any $\mathbf{x} \in \left[ 0, 1\right]^d$.
\item[(A3)] The eigenvalues of $G(\cdot,\cdot)$ satisfy $\lambda_1 \geq \lambda_2 \geq \ldots \geq 0$ and  $\sum_{k=1}^{\infty}\lambda_k < \infty$. Moreover, the maximal eigenvalue of the kernel matrix $\mathbf{C}_{N}$ satisfies  {$\lambda_{1,N}=O(N^{-\varrho})$} for some constant $\varrho\geq 0$.
\item[(A4)]  The DNN estimator $\widehat{f} \in \mathcal{F}(L, \mathbf{p}, s, F)$, where  $L \asymp \log (nN^{\varrho})$, $s \asymp (nN^{\varrho})^{\frac{1}{\theta+1}}$, $F \geq \max(\|\mathbf{K}\|_{\infty},1)$, $\min_{l=1, \ldots, L}p_l \asymp  (nN^{\varrho})^{\frac{1}{\theta+1}}$, for
$\theta = \min_{i=0,\ldots,q}\frac{2\beta_i^{\ast}}{t_i}$.
\end{description}

Assumption (A1)  is a natural definition for neural network, which is fairly flexible and many well known function classes are contained in it. For example, the  additive model
 $f_0(\mathbf{x}) = \sum_{i=1}^df_i(x_i)$,
   can be written as a composition of two functions
 $f_0 = g_1 \circ g_0$,
  with $g_0(\mathbf{x}) = \left( f_1(x_1), \ldots, f_d(x_d)\right)^{\top}$ and $g_1(\mathbf{x}) = \sum_{j=1}^d x_j$, such that $g_0 : \left[ 0,1 \right]^d \rightarrow \mathbb{R}^d$ and $g_1 :  \mathbb{R}^d \rightarrow \mathbb{R}$.  Here $\mathbf{d} = (d, d, 1)$ and $\mathbf{t} = (1, d, 1)$. The generalized additive model
   $f_0(\mathbf{x}) = h\left(\sum_{i=1}^df_i(x_i)\right)$,
 it can be written as a composition of three functions
 $f_0 = g_2\circ g_1 \circ g_0$, with $g_0$, $g_1$ described above, and $g_2 = h$.

Assumption (A2) is a standard assumption for the variance of measurement errors.  which requires the bounded variance of measurement error over the whole space. This assumption is widely used in functional data nonparametric regression  literature, see \cite{Cao:Yang:Todem:12,Yao:etal:05b} for example.  In out article, the variance function is used in Lemma 3 in the supplementary material, which bounds the first largest eigenvalue of measurement error covariance function by a constant.
Assumption (A3) is a standard eigenvalue assumption for Mercer kernel and it has been widely used assumption for covariance functions in FDA  literature, see \cite{Cao:Wang:Li:Yang:14, Li:Hsing:10a} for example.
By \cite{Braun:06}, Assumption (A3) trivially holds for $\varrho= 0$ (see Lemmas 5 and 6),
and may even hold for some positive $\varrho$ as revealed by Examples 1 in Section \ref{SEC:example1}.
Assumption (A4) depicts the architecture and parameters' setting  in the network space. 

The following theorem establishes the convergence rate of the DNN estimator $\widehat{f}$ under the empirical norm. Its proof and some technical lemmas will be provided in the Supplementary material.
\begin{theorem} \label{THM: rate}
Under Assumptions (A1)-(A4), with probability greater than $(1-\frac{2}{nN^{\varrho}})^{\ceil{\log (nN^{\varrho})}+1} \to 1$, we have
\begin{equation}\label{EQ: convergence rate}
\|\widehat{f} - f_0 \|_N^2 \leq c(nN^{\varrho})^{-\frac{\theta}{\theta+1}}\log^6 (nN^{\varrho}),
\end{equation}
where  ${\varrho} \geq 0$, $\theta = \min_{i=0,\ldots,q}\frac{2\beta_i^{\ast}}{t_i}$, $c$ is a constant only depends on $\mathbf{t}$, $\mathbf{d}$, $\bs{\beta}$ given in (\ref{EQ:gfunction}).
\end{theorem}

Recall that $\varrho\ge0$ characterizes the decay rate of the maximal eigenvalue of the covariance matrix $\mathbf{G}_N$.
Let $\min_{i=0,\ldots,q}\frac{2\beta_i^{\ast}}{t_i}= \frac{2 {\beta}^{\ast}}{ {t}^{\ast}}$. When $\varrho = 0$, the convergence rate  $n^{-\frac{2\beta^{\ast}}{2\beta^{\ast}+t^{\ast}}}$ is obtained  (up to $\log n$ factors), which is free of dimension $d$.
 When $\varrho > 0$ and $N\gg n^{\frac{1}{\theta \varrho}}$,
the convergence rate of the neural network is faster than $n^{-1}$.
Such a ``super-convergence'' phenomenon was first discovered by \cite{CY:11} who showed that smoothing spline estimator achieves estimation rate $n^{-1}$ under $L^2$-norm
when sampling frequency is sufficiently large. Our contribution is to rediscover this phenomenon for neural network estimator.

In the following, to explicitly demonstrate the convergence rates discussed in Theorem \ref{THM: rate} are achievable, two examples are provided. 

\subsection{Example 1}\label{SEC:example1}
Let $\mathbf{X}_j=({j_1}/{N_d},\ldots, j_d/N_d)$,
$1 \leq j_k \leq N_d$ for $k=1, \ldots, d$, be the evenly spaced grid points of $[0,1]^d$,
where $N_d = N^{1/d}$, $d\geq 1$.  Consider a Bernoulli polynomial kernel function
$
G_0(x,x') = 2\sum_{k=1}^{\infty}\frac{\cos(2\pi k(x-x'))}{(2\pi k)^{\varrho d}}$, $x,x'\in[0,1]$,
where $\varrho > 1$. See \cite{Wahba:90} for an introduction of such kernel.
For $k=1,\ldots,d$,
the kernel matrix on the $k$-th coordinate of $\mathbf{X}_j$ is $\mathbf{C}_{N, k} = \left\{ N^{-1}G_0(j_k/N_d, j'_k/N_d)\right\}_{j_k,j'_k=1}^{N_d}$.
Assume that the covariance matrix $N\mathbf{C}_{N}$ has an
additive structure such that $\mathbf{C}_{N} =\sum_{k=1}^d\mathbf{C}_{N,k} $.
We require certain order of grid points by sorting them  based on the $d$-th coordinate values first, then by the $(d-1)$-th coordinate values, and so on, until
we reach the first coordinate. Let $\mathbf{A}_{N_d }$
be an $N_d \times N_d$ matrix whose $(\ell, \ell')$-th entry is
$2N_d^{-1}\sum_{k=1}^{\infty}\frac{\cos(2\pi k(\ell-\ell')/N_d)}{(2\pi k)^{\varrho d}}$, $\ell, \ell'=1,\ldots,N_d$, and  $\mathbf{1}_{N_d }$ be the all-ones $N_d \times N_d$ matrix. Then we have the following relationship:
\begin{eqnarray*}
&&\mathbf{C}_{N,1} = N_d^{1-d}\times\mathbf{1}_{N_d }\otimes \mathbf{1}_{N_d }\otimes \ldots \otimes \mathbf{1}_{N_d } \otimes \mathbf{A}_{N_d }, \\
&&\mathbf{C}_{N,2} = N_d^{1-d}\times\mathbf{1}_{N_d }\otimes \mathbf{1}_{N_d }\otimes \ldots \otimes\mathbf{A}_{N_d } \otimes \mathbf{1}_{N_d }, \\
&&\ldots  \ldots ,\\
&&\mathbf{C}_{N,d-1} = N_d^{1-d}\times\mathbf{1}_{N_d }\otimes \mathbf{A}_{N_d } \otimes \ldots \otimes \mathbf{1}_{N_d }\otimes \mathbf{1}_{N_d },\\
&&\mathbf{C}_{N,d} = N_d^{1-d}\times\mathbf{A}_{N_d }\otimes\mathbf{1}_{N_d }\otimes \ldots \otimes \mathbf{1}_{N_d }\otimes \mathbf{1}_{N_d },
\end{eqnarray*}
where $\otimes$ is the kronecker product operator. According to equation (20) in \cite{Shang:Cheng:17}, $\mathbf{A}_{N_d }$ is a circulant matrix whose eigenvalues
have an explicit expression:
\[ \lambda^{\ast}_{j} = \begin{cases}
      \hspace{2.5cm} 2\sum_{k=1}^{\infty}\frac{1}{(2\pi kN_d)^{ \varrho d}}, & j=0\\
      \sum_{k=1}^{\infty}\frac{1}{\left[ 2\pi (kN_d - j)\right]^{\varrho d}}  +  \sum_{k=0}^{\infty}\frac{1}{\left[ 2\pi (kN_d + j)\right]^{\varrho d}},  & 1\leq j \leq N_d-1 .
   \end{cases}
\]
In  the Appendix C, we have shown that $\max_{j=1,\ldots, N_d}\lambda_j^{\ast} \lesssim N_d^{- \varrho d}$.
Since the maximal eigenvalue  of $\mathbf{1}_{N_d }$  is $N_d$, by the property of Kronecker product, the maximal eigenvalue of $\mathbf{C}_{N,k}$ is $O(N^{-\varrho})$.
 Consequently, the first largest eigenvalue for $\mathbf{C}_{N}$ is $\lambda_{1, N} \lesssim  N^{-\varrho} $.
  According to Assumption (A3), this ensures the better convergence rate in equation (\ref{EQ: convergence rate}). When {$N \gg n^{\frac{1}{\varrho\theta}}$}, the convergence rate of $\widehat{f}$ is faster than $n^{-1}$.

\subsection{Example 2}

Define  a cosine random process
$
\Lambda_k(2\pi x) = \xi_k \cos(2\pi x) + \xi_k ' \sin(2\pi x)$,
where $\xi_k$ and $\xi_{k}'$ are identically distributed and uncorrelated, with mean zero and covariance $\mathrm{E} \xi^2$.
 According to \cite{Taylor:09}, the covariance function for cosine process is given by
\[
G_{0}\left(j_k/N_d,j'_k/N_d\right)= \mathrm{E}{\xi^2} \cos\left(2\pi \left(j_k - j'_k\right)/N_d\right)
\]
and
\[
G_{0}(\mathbf{X}_j, \mathbf{X}_{j'}) = d^{-1}\mathrm{E} \xi^2\sum_{k=1}^d  \cos\left(2\pi \left(j_k - j'_k\right)/N_d\right),
\]
which is the $(j, j')$-th entry in covariance matrix $\mathbf{C}_{N}$.

Therefore, $\mathbf{C}_{N}$ can be written as  $\mathbf{C}_{N} = \sum_{k=1}^d  \mathbf{C}_{N,k}$, where $\mathbf{C}_{N,k}$ is the
kernel matrix for the $k$-th coordinate of $\mathbf{X}_j$, with $(j, j')$-th entry $N^{-1}\cos\left(2\pi \left(j_k - j'_k\right)/N_d\right)$.
Let $\mathbf{B}_{N_d }$ be an $N_d \times N_d$ matrix whose $(\ell,\ell')$-th entry is $N_d^{-1}\cos\left(2\pi \left(\ell - \ell'\right)/N_d\right)$, $\ell,\ell'=1,\ldots,N_d$. Similar to Example 1, we require the certain order of the grid points and thus have the following relationship:
\begin{eqnarray*}
&&\mathbf{C}_{N,1} = N_d^{1-d}\times\mathbf{1}_{N_d }\otimes \mathbf{1}_{N_d }\otimes \ldots \otimes \mathbf{1}_{N_d } \otimes \mathbf{B}_{N_d }, \\
&&\mathbf{C}_{N,2} = N_d^{1-d}\times\mathbf{1}_{N_d }\otimes \mathbf{1}_{N_d }\otimes \ldots \otimes\mathbf{B}_{N_d } \otimes \mathbf{1}_{N_d }, \\
&&\ldots  \ldots ,\\
&&\mathbf{C}_{N,d-1} = N_d^{1-d}\times\mathbf{1}_{N_d }\otimes \mathbf{B}_{N_d } \otimes \ldots \otimes \mathbf{1}_{N_d }\otimes \mathbf{1}_{N_d },\\
&&\mathbf{C}_{N,d} = N_d^{1-d}\times\mathbf{B}_{N_d }\otimes\mathbf{1}_{N_d }\otimes \ldots \otimes \mathbf{1}_{N_d }\otimes \mathbf{1}_{N_d },
\end{eqnarray*}

Since $\mathbf{B}_{N_d }$ is a circulant matrix, its maximal eigenvalue $\lambda_1^{\ast}$ can be explicitly written as  $N_d^{-1}\sum_{k=0}^{N_d-1}\cos\left(2\pi k/N_d  \right) \omega ^{N_d - k}$, where $\omega = \exp\left( 2\pi \sqrt{-1}/N_d \right)$. By direct calculations,
it can be shown that $\lambda_1^{\ast} = N_d/2$.
Since the maximal eigenvalue  of $N_d^{-1}\mathbf{1}_{N_d }$  is $1$, by the property of Kronecker product, it follows that the maximal eigenvalue of $\mathbf{C}_{N,k}$ is $1/2$.
 Consequently, the maximal eigenvalue of $\mathbf{C}_{N}$ is $\lambda_{1, N} \asymp \mathrm{E} (\xi^2)/2=O(1)$.
 According to the trivial case ($\varrho = 0$) in Assumption (A3), we have the usual nonparametric convergence rate for $\|\widehat{f} - f_0\|^2_N$ as $O(n^{-\frac{\theta}{\theta+1}}\log^6n)$.

\section{Simulation}
\label{SEC:sim}

To illustrate how the introduced nonparametric
regression estimators based on our proposed
neural networks method behave in case of finite sample sizes, we  conduct substantial simulations for both 2D  and 3D  functional data.

 \subsection{2D simulation}
\begin{figure}
\begin{center}
\hspace{1.cm}\textbf{Case I } \hspace{5.5cm}\textbf{Case II}\\
\textbf{$f_0$}
$\begin{array}{l}
\includegraphics[trim = 30mm 0mm 30mm 0mm, clip,width = .4\textwidth]{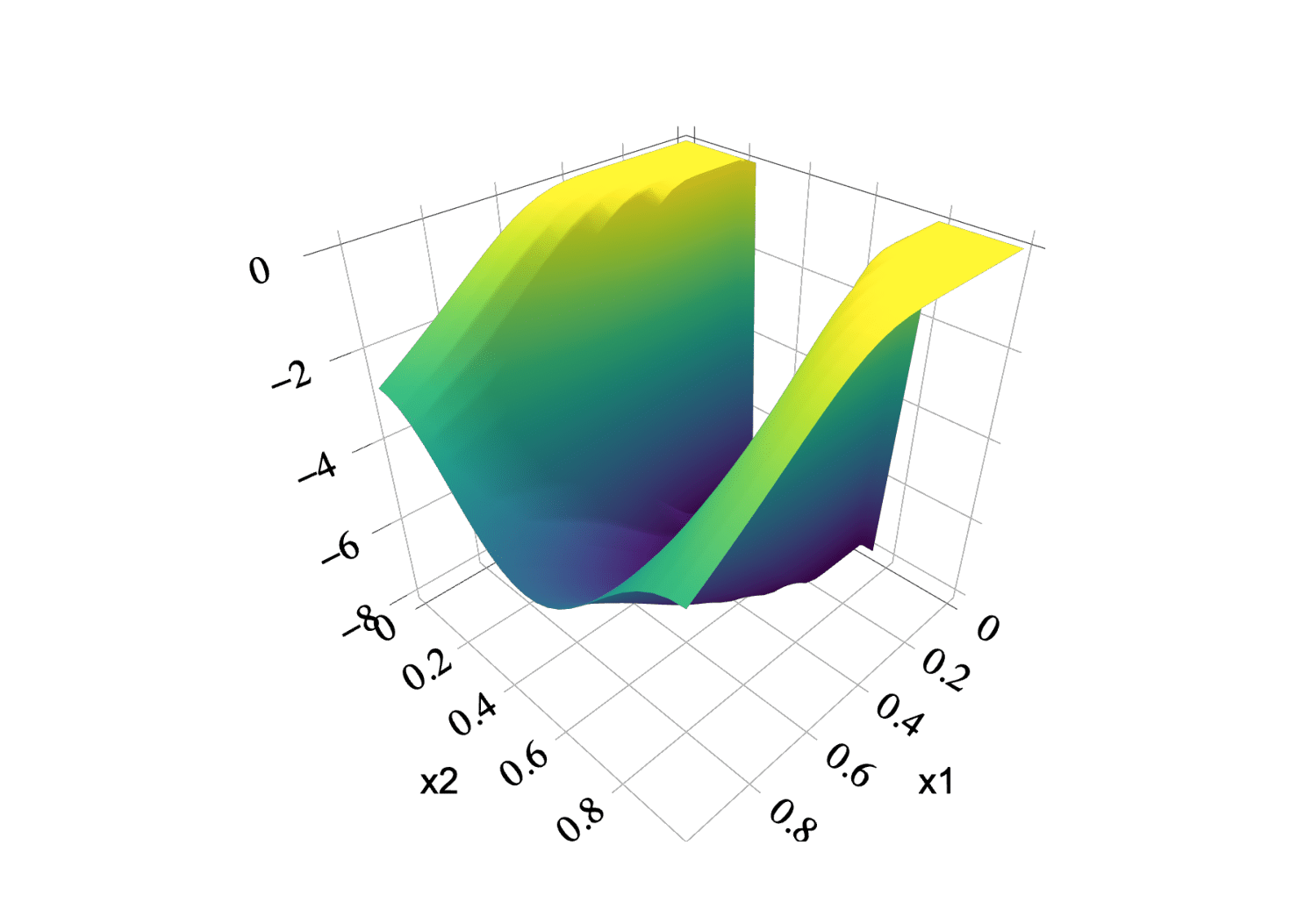}  
\includegraphics[trim = 30mm 0mm 30mm 0mm, clip,width = .4\textwidth]{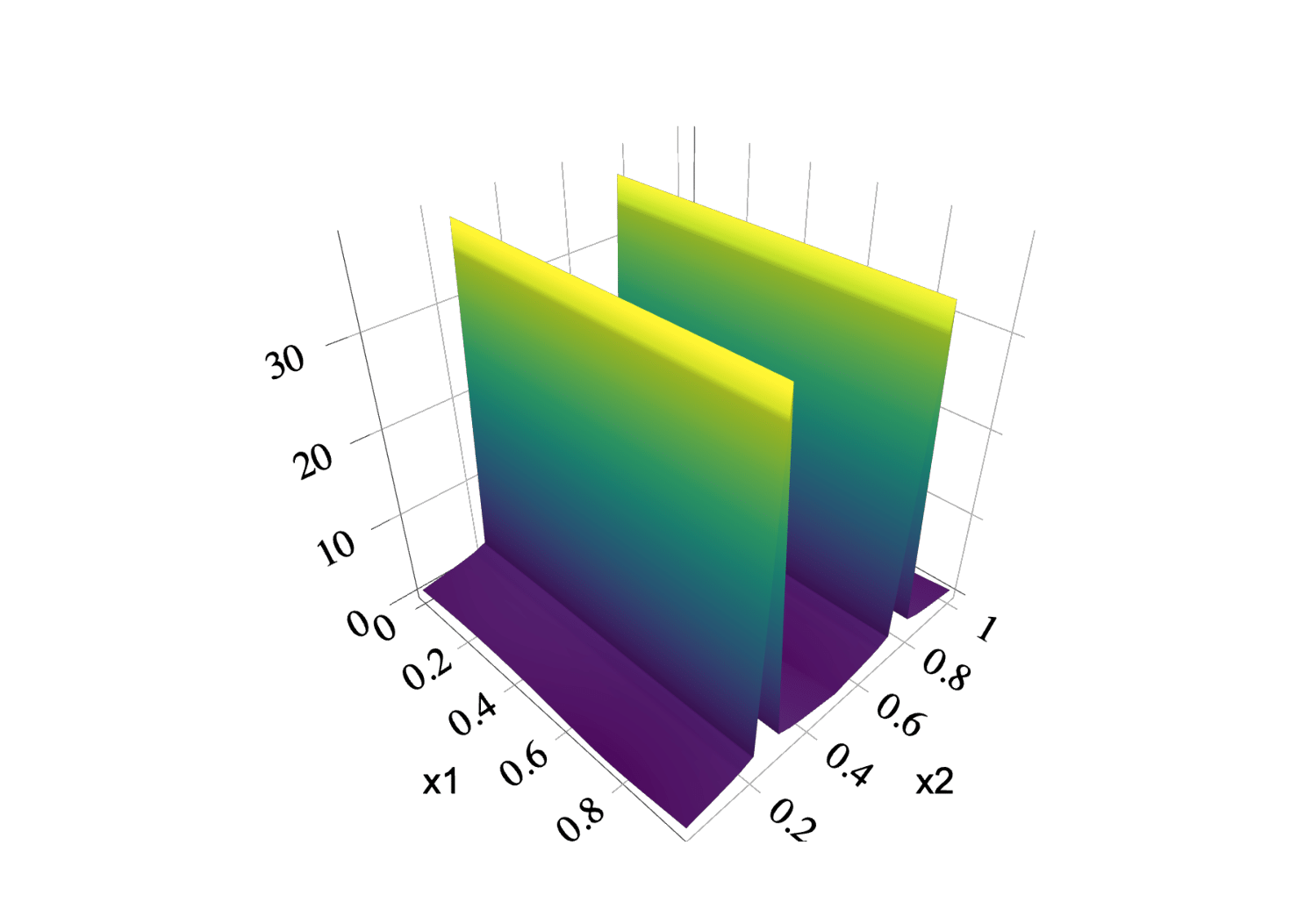} \\
\end{array}$\\
\textbf{$\hat{f}_{DNN}$}
$\begin{array}{l}
\includegraphics[trim = 30mm 0mm 30mm 0mm, clip,width = .4\textwidth]{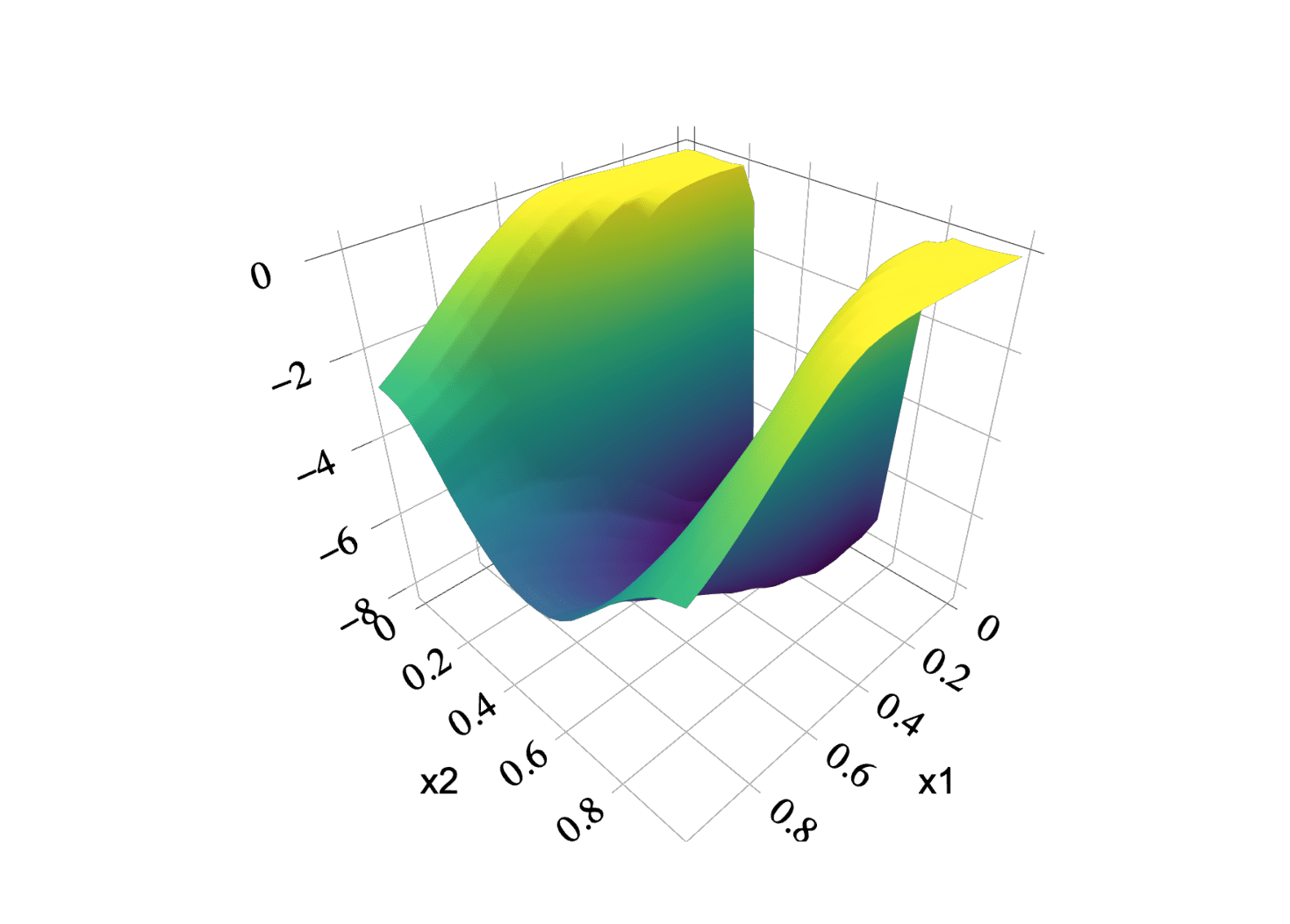}  
\includegraphics[trim = 30mm 0mm 30mm 0mm, clip,width = .4\textwidth]{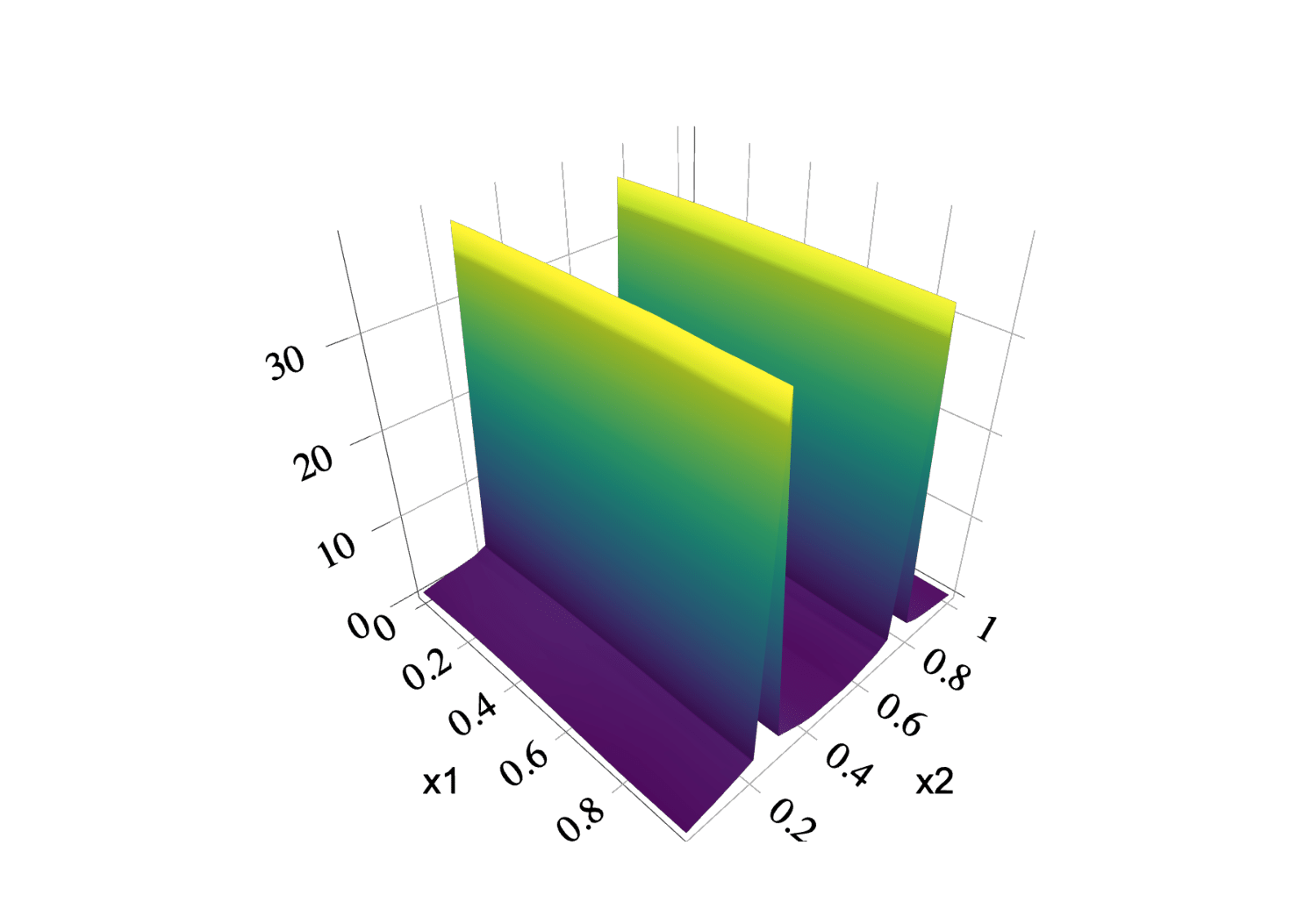} \\
\end{array}$\\
\textbf{$\hat{f}_{BS}$}
$\begin{array}{l}
\includegraphics[trim = 30mm 0mm 30mm 0mm, clip,width = .4\textwidth]{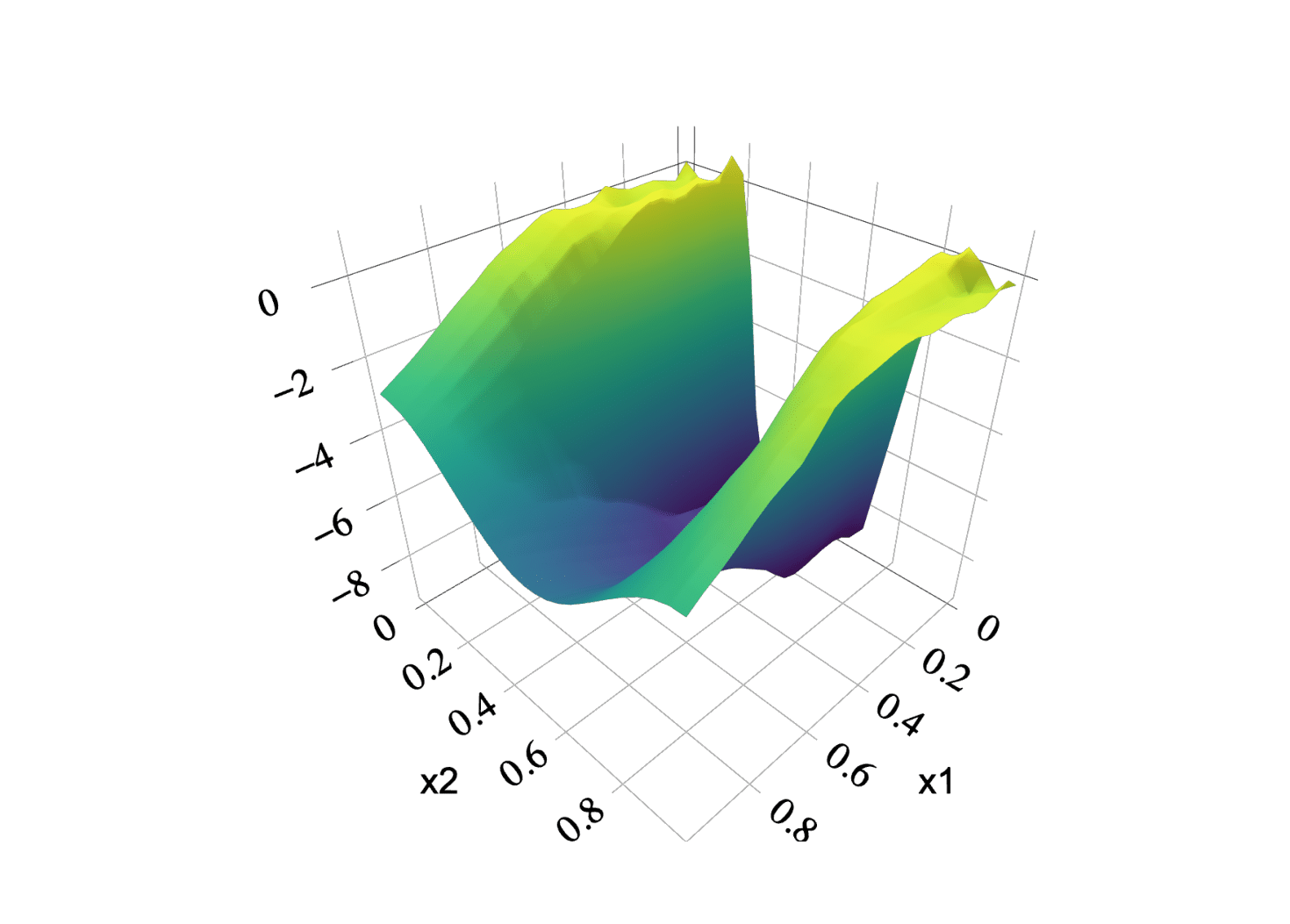}  
\includegraphics[trim = 30mm 0mm 30mm 0mm, clip,width = .4\textwidth]{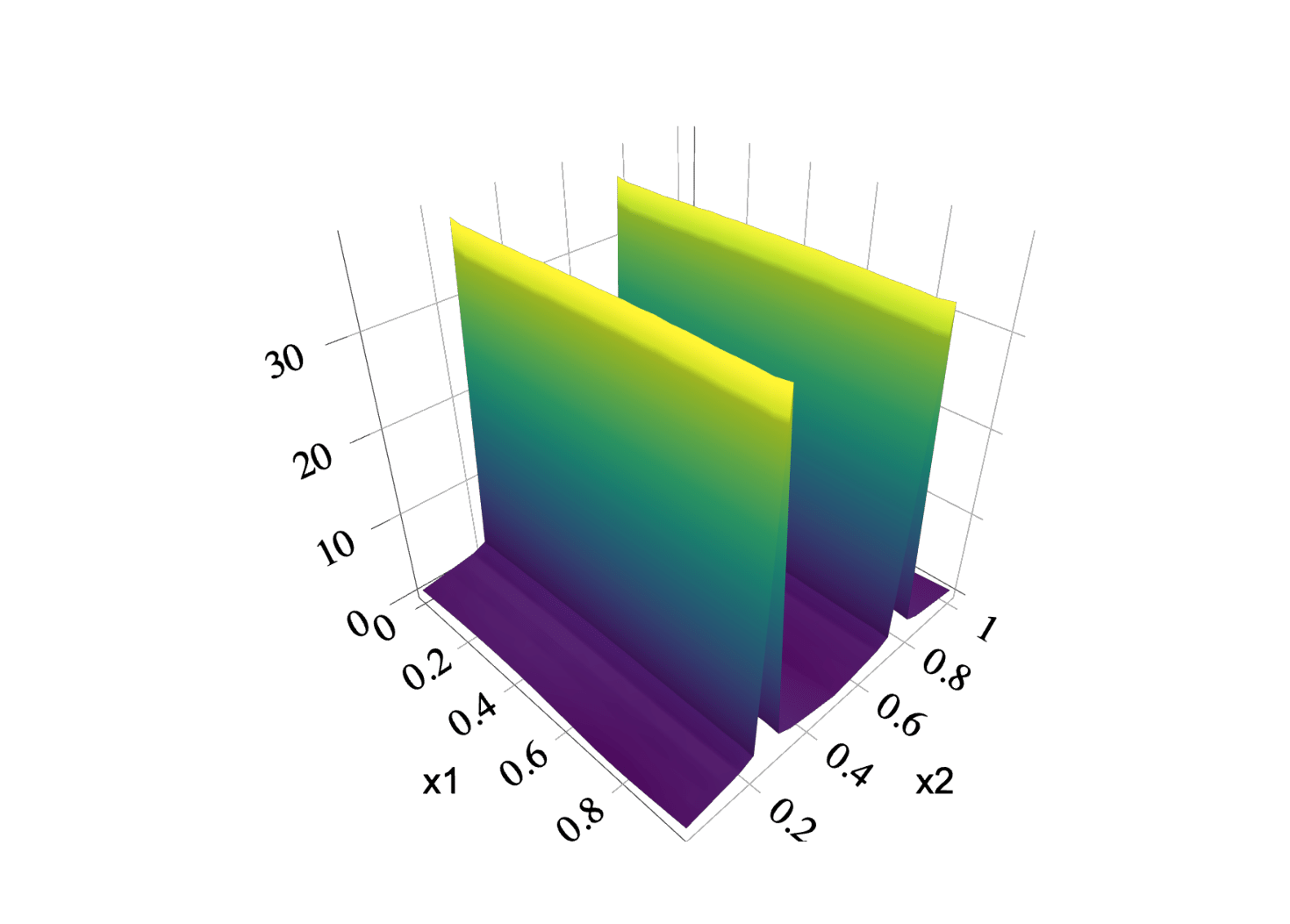} \\
\end{array}$
\caption{2D simulation. Left: from the top to bottom, they are true function $f_0$ (Case 1) and its estimators $\widehat{f}_{DNN}$ and $\widehat{f}_{BS}$.
Right: from the top to bottom, they are true function $f_0$ (Case 2) and its estimators $\widehat{f}_{DNN}$ and $\widehat{f}_{BS}$.  ($n=200$, $N=625$ and $\sigma=1$)}
\label{FIG: simulation 2d}
\end{center}
\end{figure}

 In this simulation, the 2D images are generated from the model:
\begin{equation} \label{EQ:sim}
Y_{ij} =f_0\left(\mathbf{X}_j\right) + \eta_i\left(\mathbf{X}_j\right) + \epsilon_{i}\left(\mathbf{X}_j\right),
\end{equation}
where $\mathbf{X}_j=(X_{1j}, X_{2j})= \left(j_1/N_2, j_2/ N_2 \right)$, $1 \leq j_1, j_2 \leq N_2$ are equally spaced grid points on the $\left[ 0, 1\right] ^2$ and $N_2^2=N$.
To demonstrate the practical performance of our
theoretical results, we consider the following two mean functions:
\begin{itemize}
\item Case 1  :  $f_0(x_{1j},x_{2j})=\frac{-8}{1+\exp\left( \cot(x_{1j}^2)\cos(2\pi x_{2j})\right)}$,
\item Case 2  : $f_0(x_{1j},x_{2j})=\log\left( \sin(2\pi x_{1j}) + 2|\tan(2\pi x_{2j})| + 2 \right)$,
\end{itemize}
and the corresponding images are shown in the first row of Figure \ref{FIG: simulation 2d}.
To simulate the within-subject dependence for each subject $i$, we generate   $\eta_i\left( \cdot \right)$  from a Gaussian process, with mean $0$, and covariance function $
G_{0}\left(\mathbf{x}_{j},\mathbf{x}_{j'} \right) = \sum_{k=1}^2\cos\left(2\pi (x_{kj}-x_{kj'}) \right)$, $j,j'=1,\ldots,N$.  We generate $\epsilon_{i}\left(\mathbf{x}_j\right)=\varepsilon_{ij}\sim^{\mbox{i.i.d.}} \mathcal{N}(0,\sigma^2)$ for $i=1,\ldots,n$, $j=1,\ldots,N$. The noise level is set to be $\sigma=1,2$. We consider sample size $n= 50, 100, 200$ and for each  image, let $N_2  = 15$ or  $25$, which means for each 2D image, the number of observational points (pixels) is set to be $N=N_2^2 = 225$ or $625$. 

The parameters $L$ and $\mathbf{p}$ which represent the depth and the width of the neural network, are chosen in a data-dependent way. After some prior work of tuning parameters   based on  Assumption (A4), we use $3$ hidden layers, and different neuron numbers based on different settings. In practice, we set the same neuron numbers for each layer for simplicity, and we follow the rule that the neural numbers are increasing as $n$ and $N$ are increasing.  Sparsity parameters $s$ is intrinsically determined by the built in $L_1$ penalty in R package \texttt{keras}.  The batch size is a hyper-parameter that defines the number of samples to work through before updating the internal model parameters. We choose $32$ or  $64$ batch size depending on the performance of convergence. The number of epochs is also a hyper-parameter which defines the number times that the learning algorithm works through the entire training data set. We select $300$ or $500$ epochs to obtain the convergent results depending on different cases as well. In our settings, we recommend optimizer Adam (adaptive moment estimation). Adam is an optimization algorithm that can be used instead of the classical stochastic gradient descent procedure to update network weights iterative based in training data  (see \cite{Kingma:Ba:15}). There are several other state-of-art  gradient-based optimization algorithms, such as  stochastic gradient decesendant and Adam.   We have applied these optimization algorithms, and Adam provides the best results and is the most computationally efficient in our simulation study among these candidates.

The alternative approach for 2D case we considered is a 2D regression spline method (bivariate spline). With regard to the variety of modifications of this approach
known in the literature, we focus on the version for 2D FDA in \cite{Wang:Wang:Wang:Ogden:19}.  Let $\mathbf{B}^{\top}(\mathbf{x})=\{B_{m}(\mathbf{x})\}_{m \in \mathcal{M}}$ be the set of bivariate Bernstein basis polynomials, where $\mathcal{M}$ stands for an index set of Bernstein basis polynomials. Then we can represent any bivariate function $f(\mathbf{x})$ by $f(\mathbf{x}) \approx \mathbf{B}^{\top}(\mathbf{x})\bs{\gamma}$ where $\bs{\gamma}^{\top} =(\gamma_{m},m \in \mathcal{M})$ is the bivariate spline coefficient vector. The estimator $\widehat{f}_{BS}$ is implemented by the R package \texttt{ImageSCC}, which was developed by  the authors of \cite{Wang:Wang:Wang:Ogden:19} and
is available from \url{https://github.com/funstatpackages/ImageSCC}.

To exam the performance of the estimator $\widehat{f}$, we present the empirical $L_2$ risk, which is defined as:
\begin{equation*}
\frac{1}{N} \sum_{j_1=1}^{N_2}\sum_{j_2=1}^{N_2} \left\{ \widehat{f}\left(j_1/N_2, j_2/N_2\right) - f_0\left(j_1/N_2, j_2/N_2\right)\right\}^2.
\end{equation*}
The second and the third rows in Figure \ref{FIG: simulation 2d} depicts the proposed neural network estimator $\widehat{f}_{DNN}$ and bivariate spline estimator $\widehat{f}_{BS}$ when $n=200$, $N=625$ and $\sigma=1$. Table \ref{TAB:MSE1} summarizes the empirical $L_2$ risk  and standard deviation of estimators $\widehat{f}_{DNN}$ and $\widehat{f}_{BS}$ under $100$ simulations for two different noise levels.   From the above figures and table, one can see that  our method and the bivariate spline method have fairly similar estimation performances. As the bivariate spline estimator is able to achieve  the optimal nonparametric convergence rate \cite{Wang:Wang:Wang:Ogden:19}, the comparable estimation results in  Table \ref{TAB:MSE1}  also support the  asymptotic convergence rate of our proposed estimator $\widehat{f}_{DNN}$  in Theorem  \ref{THM: rate}.

\begin{table}[h!]  \centering
  \caption{The average empirical $L_2$ risk and their standard deviations of $f_0$ across $100$ simulation runs (2D case).}
   \label{TAB:MSE1}
\begin{tabular}{@{\extracolsep{0.1pt}}  cccccccccc}
\\[-1.8ex]
\hline \\[-1.8ex]
\multicolumn{8}{c}{ $f_0(x_{1j},x_{2j})=\frac{-8}{1+\exp\left( \cot(x_{1j}^2)\cos(2\pi x_{2j})\right)}$}\\
\hline \\[-1.8ex]
\multirow{2}{*}{$\sigma$} &\multirow{2}{*}{$N$} & \multirow{2}{*}{$n$}  & \multicolumn{2}{c}{DNN}&& \multicolumn{2}{c}{bivariate spline}\\ \cline{4-5} \cline{7-8} \\[-1.8ex]
 & & & $L_2$ risk & SD  & & $L_2$ risk & SD\\
\hline \\[-1.8ex]
  \multirow{6}{*}{1}&& 50   & 0.1327& 0.1905 && 0.6030 & 0.0418 \\
 &225& 100    & 0.0797  & 0.1244  & &0.5757 & 0.0249 \\
& & 200     & 0.0432  & 0.0574 &&0.5584 & 0.0120\\\cline{2-8} \\[-1.8ex]
&  & 50   & 0.0770 & 0.0497 && 0.1497 & 0.0462\\
&625 & 100    & 0.0535  & 0.0368   &&0.1136 & 0.0214\\
&& 200     & 0.0352 & 0.0295    && 0.0987 & 0.0098 \\
\hline \\
\multirow{6}{*}{2}& & 50   & 0.1880& 0.1521 && 0.6564 & 0.1009 \\
&225& 100    & 0.0918  & 0.0793  & &0.6035& 0.0619 \\
&& 200     & 0.0593  & 0.0529 &&0.5765 & 0.0316\\\cline{2-8} \\[-1.8ex]
&  & 50   & 0.1594 & 0.1555 && 0.2241 & 0.1218\\
&625 & 100    & 0.0862  & 0.0755   &&0.1430 & 0.0557\\
&& 200     & 0.0420 & 0.0412    && 0.1098 & 0.0232 \\
\hline \\[-1.8ex]
\multicolumn{8}{c}{$f_0(x_{1j},x_{2j})=\log\left( \sin(2\pi x_{1j}) + 2|\tan(2\pi x_{2j})| + 2 \right)$}\\
\hline \\[-1.8ex]
\multirow{2}{*}{$\sigma$} &\multirow{2}{*}{$N$} & \multirow{2}{*}{$n$}  & \multicolumn{2}{c}{DNN}&& \multicolumn{2}{c}{bivariate spline}\\ \cline{4-5} \cline{7-8} \\[-1.8ex]
 & & & $L_2$ risk & SD  & & $L_2$ risk & SD\\
\hline \\[-1.8ex]
 \multirow{6}{*}{1}&  & 50   & 0.0731 & 0.0446 && 0.0804 & 0.0382 \\
&225& 100    & 0.0437   & 0.0249  & &0.0517 & 0.0186 \\
&& 200     & 0.0254  & 0.0217 &&0.0351& 0.0100\\\cline{2-8} \\[-1.8ex]
 & & 50   & 0.0560  & 0.0206 && 0.0751 & 0.0351\\
&625 & 100    & 0.0351  & 0.0128  &&0.0541 & 0.0254\\
&& 200     & 0.0245 & 0.0085    && 0.0383 & 0.0110 \\
\hline \\
 \multirow{6}{*}{2}& & 50   & 0.1190 & 0.0975 && 0.1290 & 0.0950 \\
&225& 100    & 0.0829  & 0.0681  & &0.0931 & 0.0597 \\
&& 200     & 0.0348  & 0.0276 &&0.0464& 0.0251\\\cline{2-8} \\[-1.8ex]
 & & 50   & 0.0573  & 0.0264 && 0.1213 & 0.0859\\
&625 & 100    & 0.0331  & 0.0132  &&0.0827 & 0.0630\\
&& 200     & 0.0139 & 0.0059    && 0.0502 & 0.0251\\
\hline \\[-1.8ex]
\end{tabular}
\end{table}

\subsection{3D simulation}
\begin{figure}
\centering
\textbf{$f_0$}
$\begin{array}{l}
\includegraphics[trim = 10mm 0mm 10mm 0mm, clip,width = 0.3\textwidth]{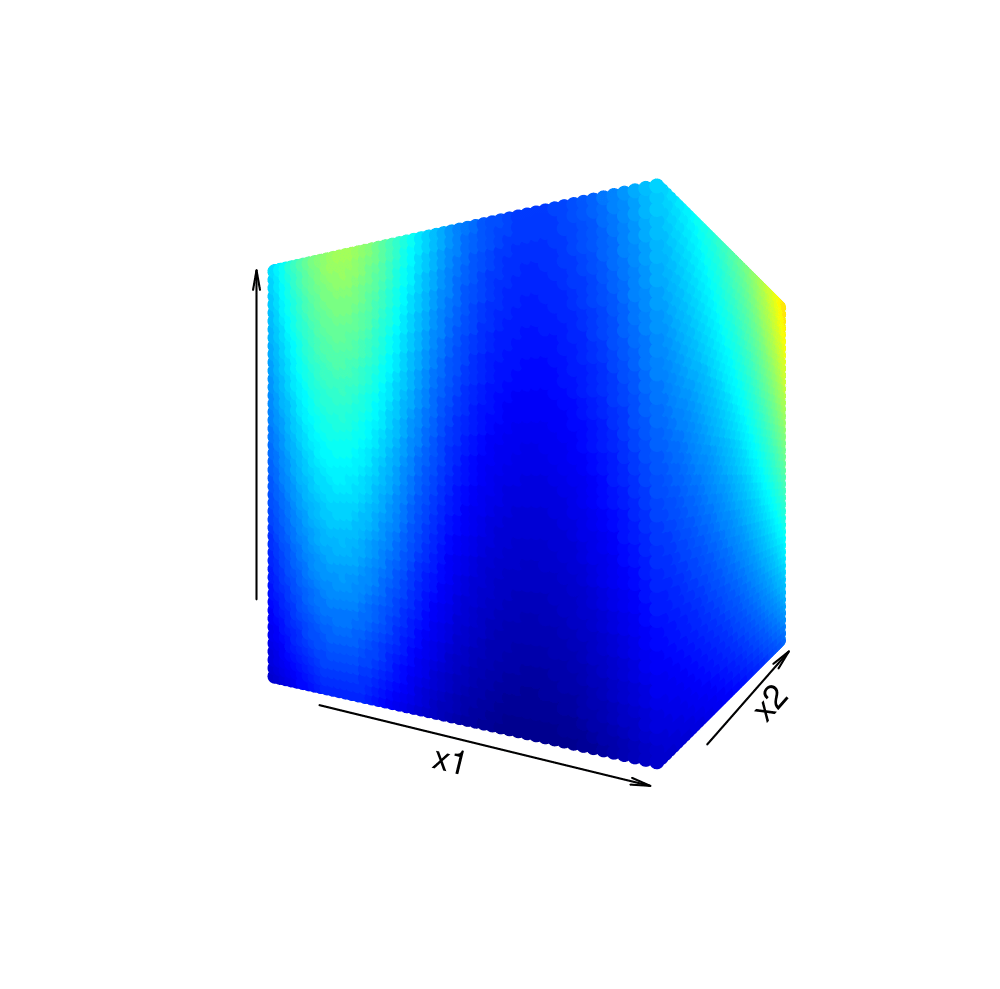}
\includegraphics[trim = 10mm 0mm 0mm 0mm, clip,width = 0.3\textwidth]{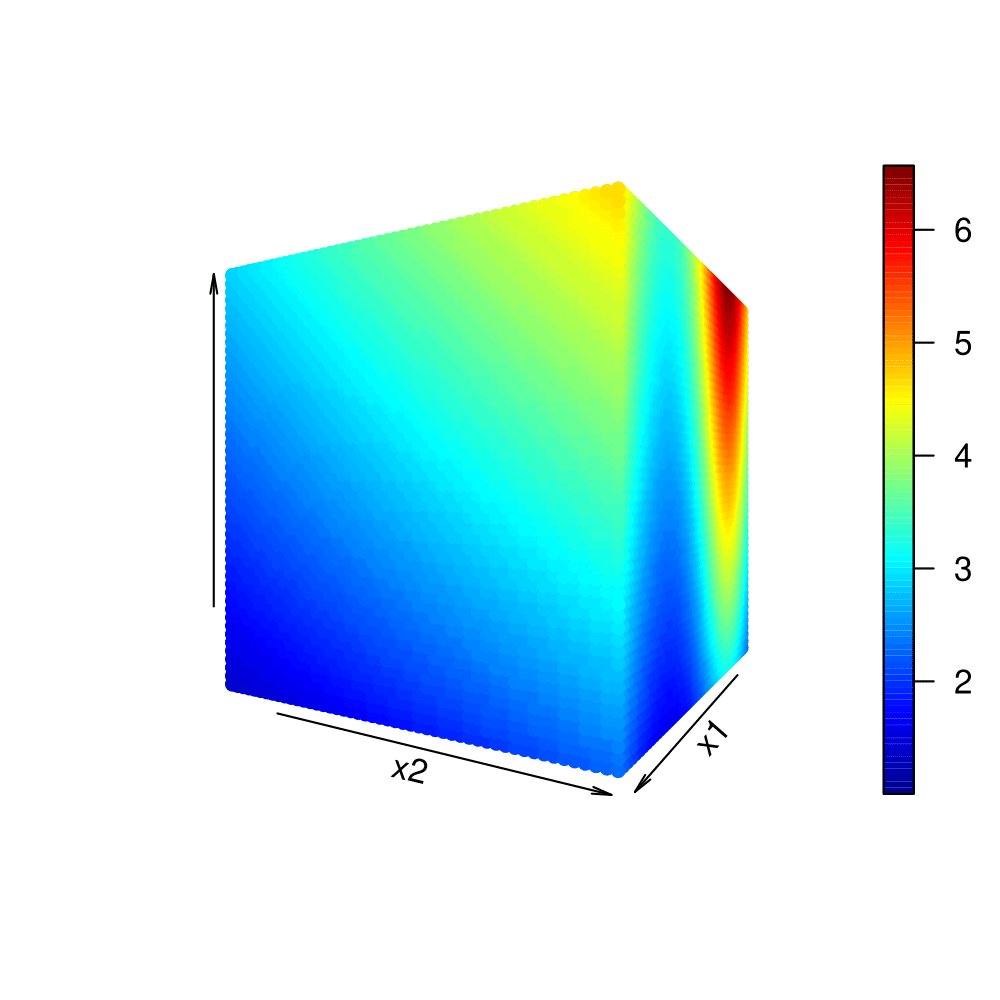}
\end{array}$\\
\vspace{-1cm}
\textbf{$\hat{f}_{DNN}$}
$\begin{array}{l}
\includegraphics[trim = 10mm 0mm 0mm 0mm, clip,width = 0.3\textwidth]{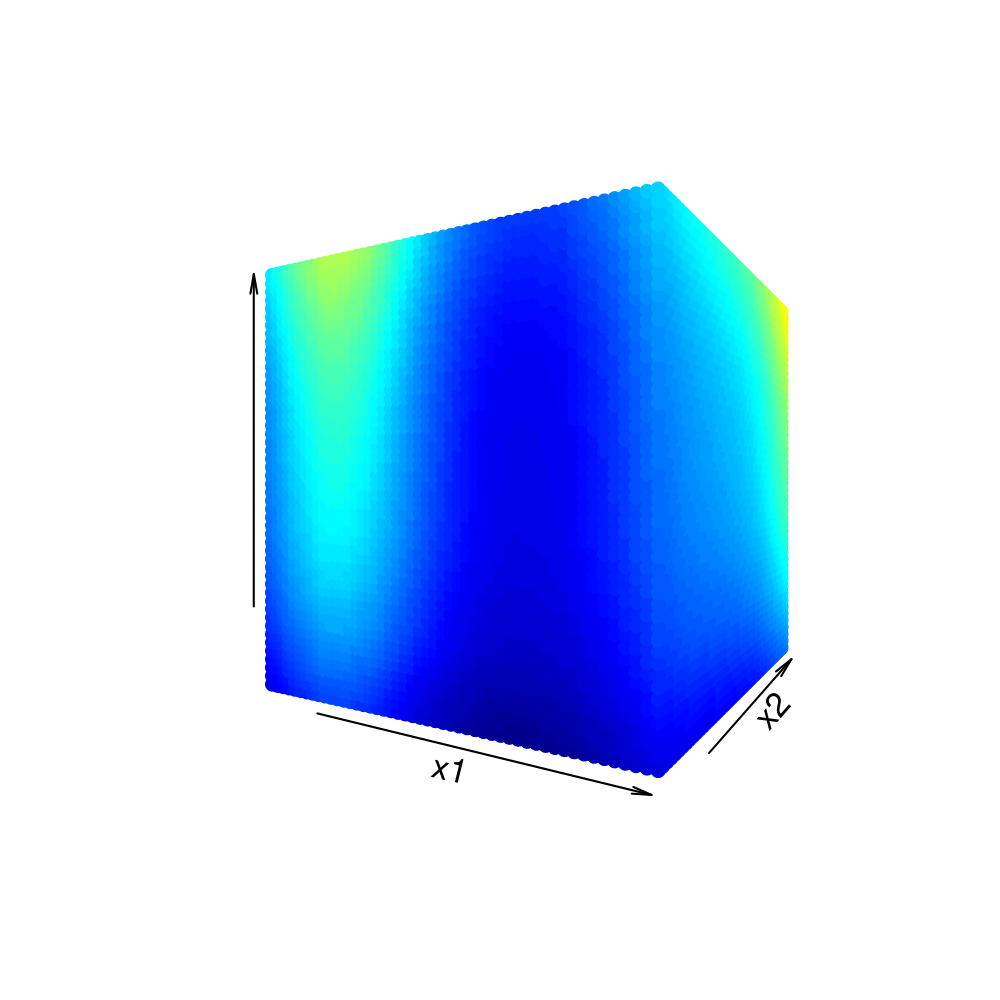}
\includegraphics[trim = 10mm 0mm 0mm 0mm, clip,width = 0.3\textwidth]{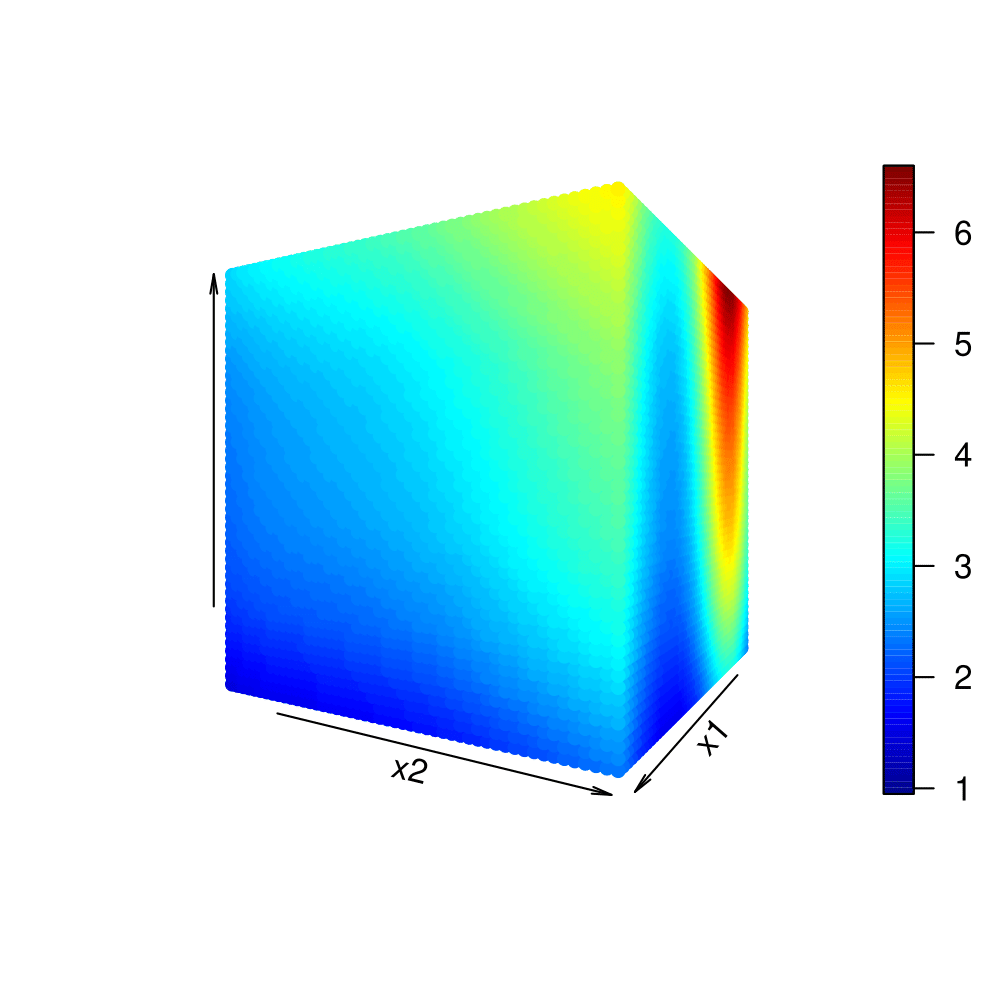}
\end{array}$\\
\caption{Two different angles (Left and Right panels) to view the true mean function and the DNN estimator in 3D simulation case. ($n=200, N=4,500, \sigma=1$) }
\label{FIG:3D}
\end{figure}

\begin{figure}
\centering
\hspace{1cm}\textbf{$f_0$} \hspace{5cm}\textbf{$\hat{f}_{DNN}$}\\
$\begin{array}{l}
\includegraphics[width = 0.4\textwidth]{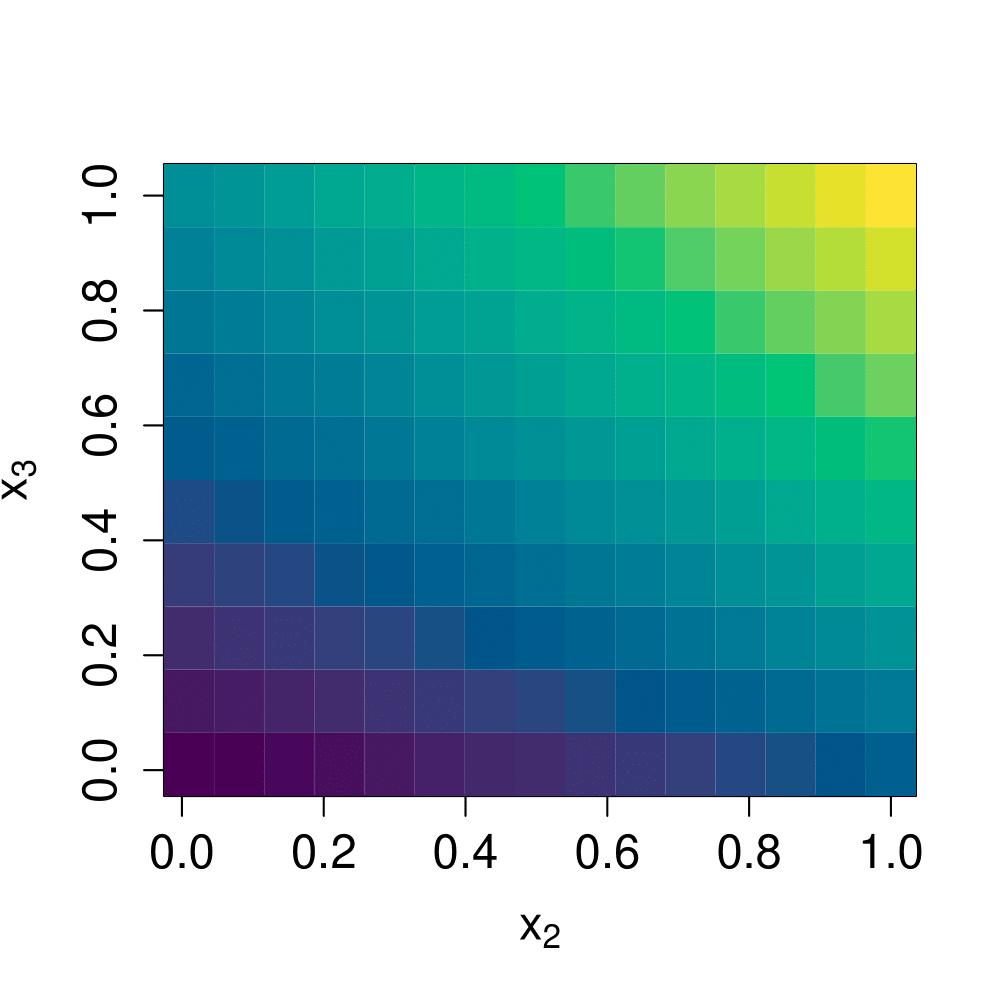}
\includegraphics[width = 0.4\textwidth]{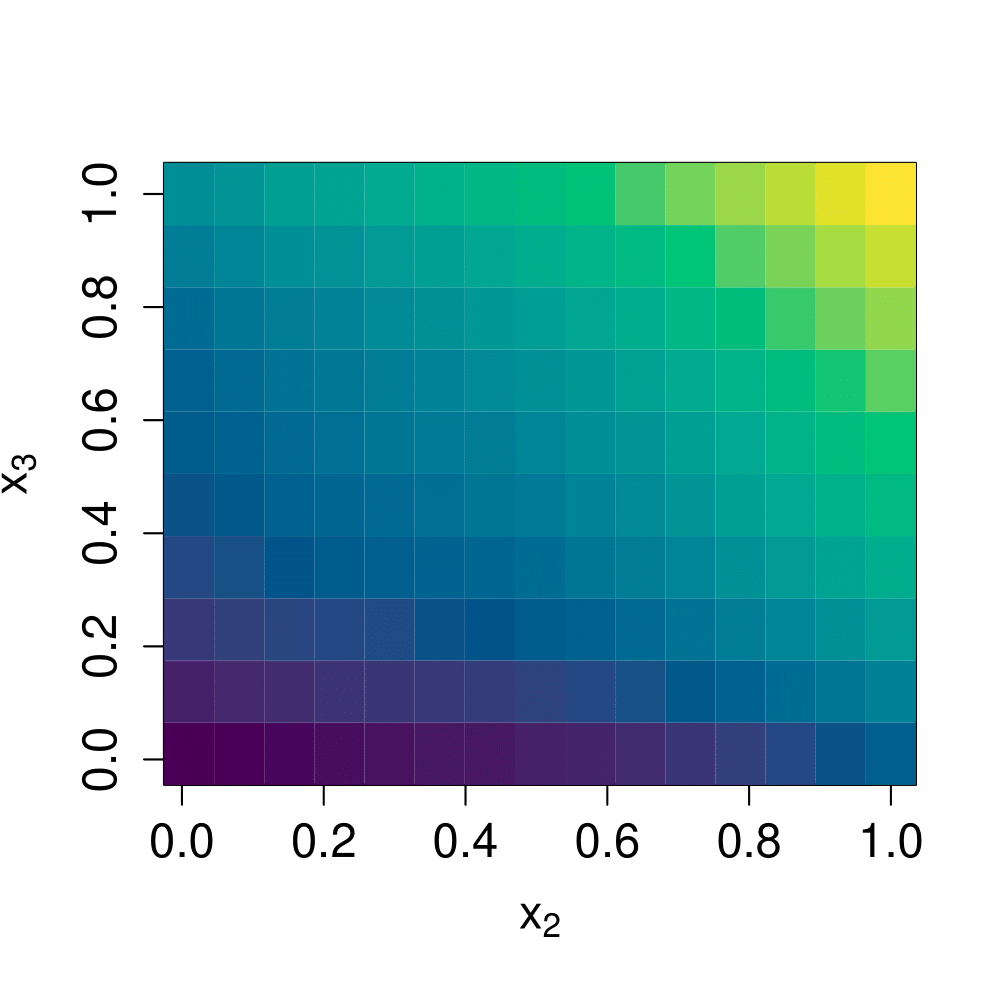} \\
\end{array}$\\
$\begin{array}{l}
\includegraphics[width = 0.4\textwidth]{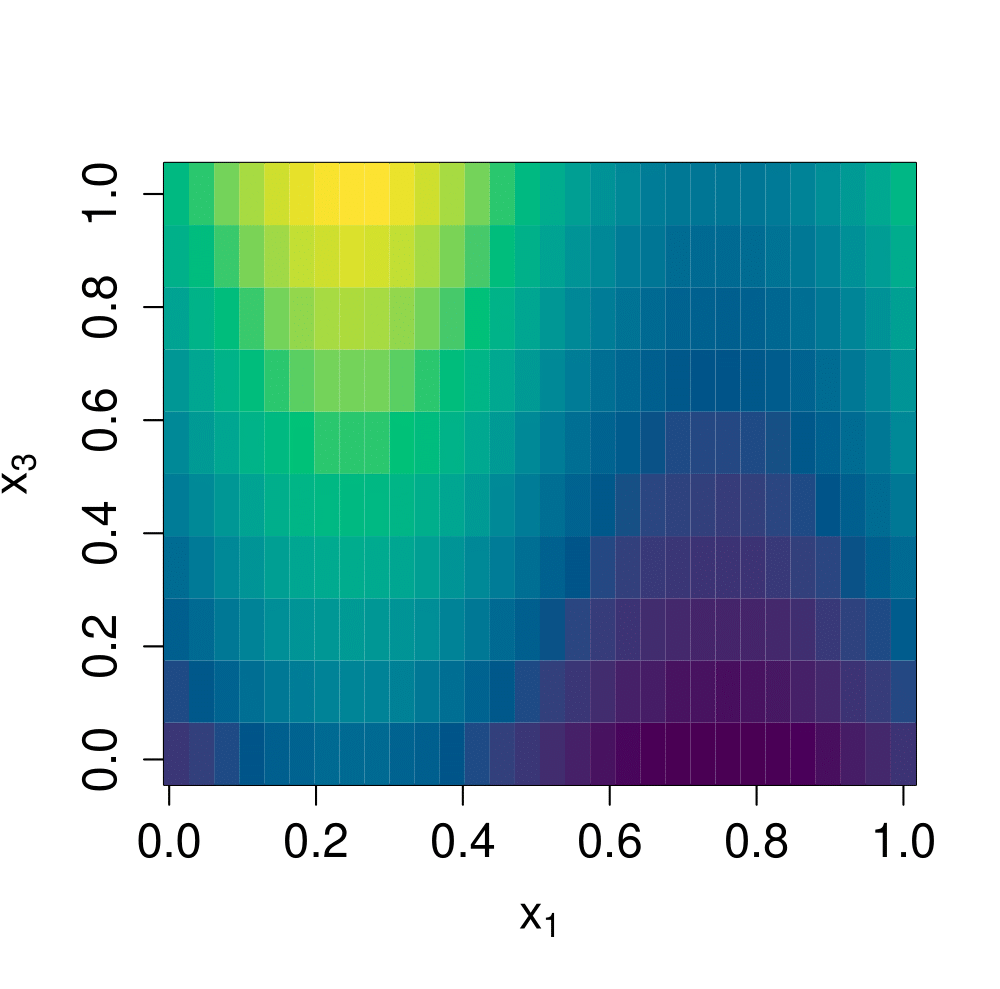}
\includegraphics[width = 0.4\textwidth]{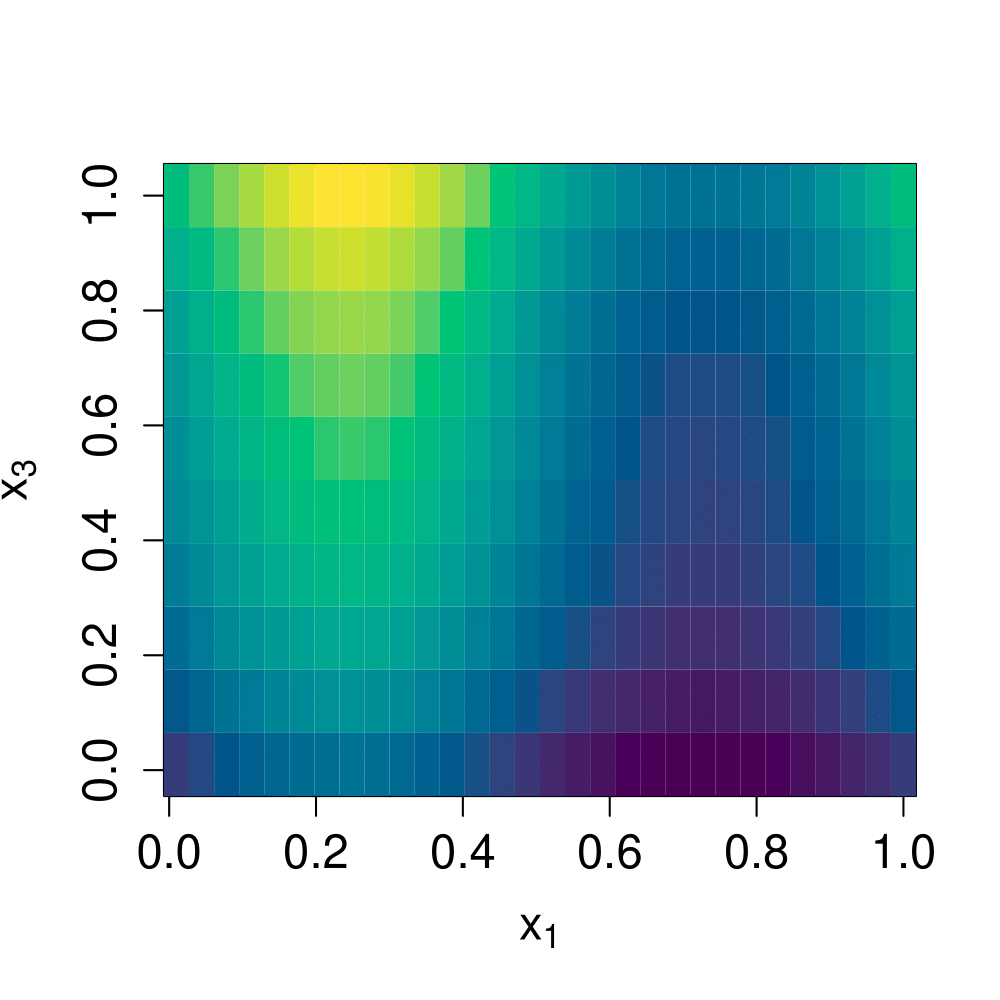} \\
\end{array}$\\
$\begin{array}{l}
\includegraphics[width = 0.4\textwidth]{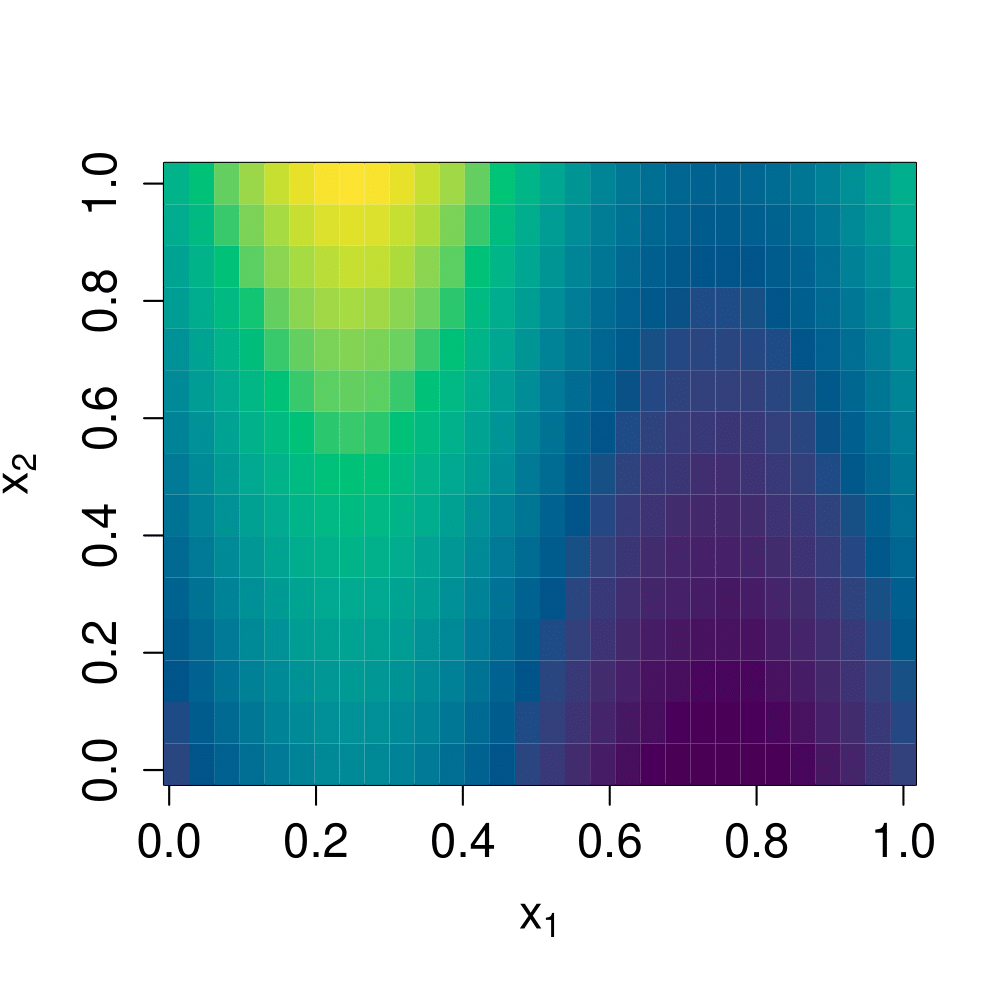}
\includegraphics[width = 0.4\textwidth]{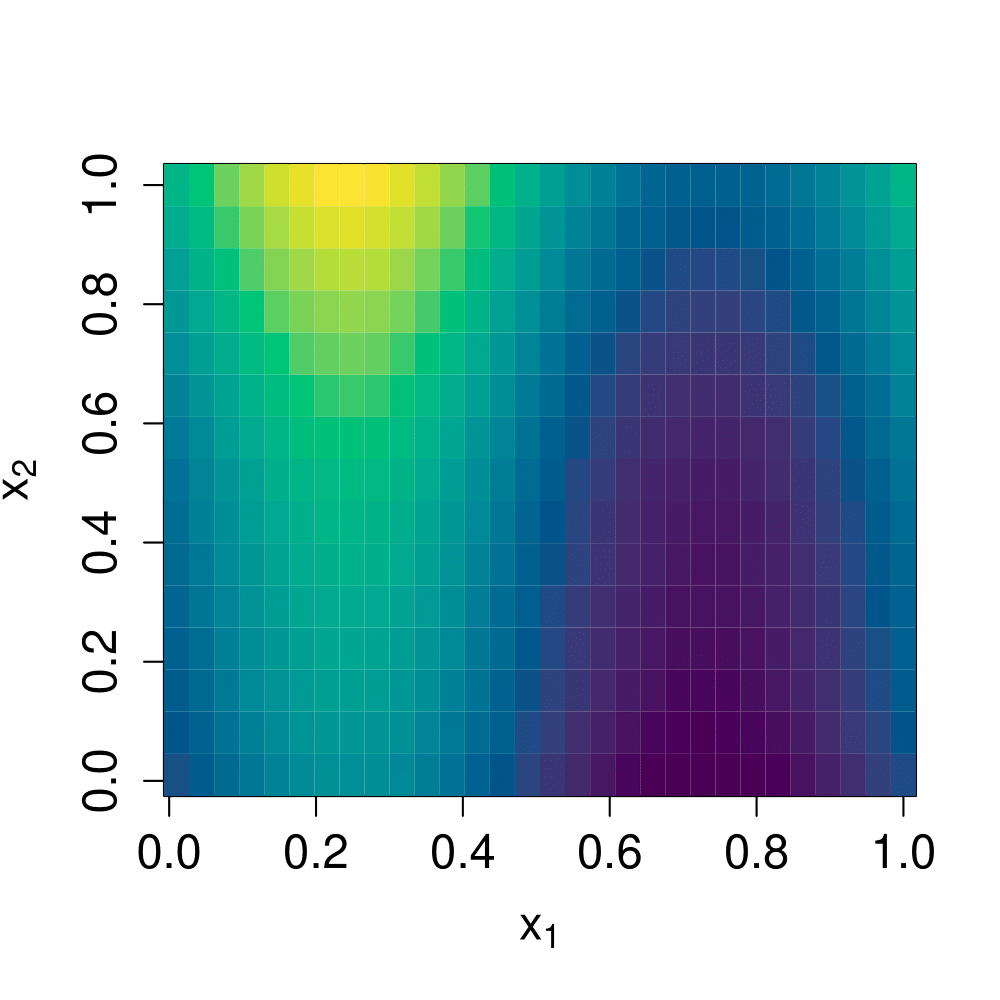} \\
\end{array}$\\
\caption{2D slices of the true mean function and the DNN estimators in 3D simulation case. Left: the true mean function $f_0$; Right: the DNN estimators. ($n=200, N=4,500, \sigma=1$) }
\label{FIG:three2Dv}
\end{figure}

For 3D simulation, the images are generated from the model (\ref{EQ:sim}) in 2D case. The true mean function is
\begin{eqnarray*}
f_0(x_{1j},x_{2j}, x_{3j})&=\exp\left(\frac{1}{3}x_{1j} + \frac{1}{3}x_{2j}+\sqrt{x_{3j}+0.1}\right)&
\end{eqnarray*}
where $(x_{1j},x_{2j},x_{3j})= \left(j_1/N_3, j_2/ N'_3,j_3/N''_3 \right)$, $1 \leq j_1\leq N_3$, $1 \leq j_2\leq N'_3$, $1 \leq j_3 \leq N''_3$ are equally spaced grid points in each dimension on $\left[ 0, 1\right]^3$ and $N_3N'_3N''_3=N$.  Here, we mimic the number of voxels of the real data, which usually have different values for $N_3$, $N'_3$ and $N''_3$. For each $i$, the within-imaging dependence $\eta_i\left( \cdot \right)$ is generated from a Gaussian process with mean $0$, and covariance function $G_{0}\left(\mathbf{x}_{j},\mathbf{x}_{j'} \right) = \sum_{k=1}^3\cos\left(2\pi (x_{kj}-x_{kj'}) \right)$, $j,j'=1,\ldots,N$. Measurement errors $\epsilon_i\left( \cdot \right)$  are generated the same as 2D case.  We consider sample size $n=50,100,200$ and $N=3,000$ ($20\times 15\times 10$) and $4,500$ ($30\times 15\times 10$). Results of each setting are based on $100$ simulations. The selection of neural network parameters follows the same rules as in 2D case.  The triangularized bivariate splines method proposed in \cite{Wang:Wang:Wang:Ogden:19} are designed for 2D functions only. Extending spline basis functions for 3D functional data  is very sophisticated and to our best knowledge, it is not available for 3D FDA yet. Hence, we only conduct 3D numerical analysis with our proposed DNN method. To exam the performance of the estimator $\widehat{f}$, we also  summarizes the empirical $L_2$ risk  and standard deviation of estimators $\widehat{f}_{DNN}$ in Table \ref{TAB:MSE3D}. It is clear to find that the empirical risk decrease when sample sizes or observed voxels numbers increase for both noise levels, which supports our theoretical findings.  The mean function $f_0$ and its DNN estimator are presented in Figure \ref{FIG:3D}. To show the detailed comparison, we also present the 2D version of   $f_0$ and its DNN estimator  in Figure \ref{FIG:three2Dv} when  $n=200$, $N=4,500$, $\sigma=1$. It is easy to conclude that the DNN estimator follows the the same pattern as the true mean function.


\begin{table}[h!]  \centering
  \caption{The average empirical $L_2$ risk and their standard deviations  of $f_0$ across $100$ simulation runs (3D case).}
   \label{TAB:MSE3D}
\begin{tabular}{@{\extracolsep{0.1pt}}  ccccc}
\\[-1.8ex]
\hline \\[-1.8ex]
 $\sigma$&$N$ & $n$& $L_2$ risk & SD \\
\hline \\[-1.8ex]
  \multirow{6}{*}{1}&& 50   & 0.0028 & 0.0020 \\
 &3000& 100    & 0.0011 & 0.0006 \\
& & 200     &0.0006&  0.0004  \\\cline{2-5} \\[-1.8ex]
&  & 50   & 0.0007 & 0.0007 \\
&4500 & 100    & 0.0005& 0.0007 \\
&& 200     & 0.0003 & 0.0004\\
\hline  \\[-1.8ex]
\multirow{6}{*}{2}& & 50   &0.0030& 0.0024  \\
&3000& 100    &0.0012 & 0.0007    \\
&& 200      & 0.0007  & 0.0005 \\\cline{2-5} \\[-1.8ex]
&  & 50   &0.0009&  0.0007 \\
&4500 & 100    & 0.0005&0.0008  \\
&& 200     & 0.0003 & 0.0005    \\
\hline \\[-1.8ex]
\end{tabular}
\end{table}

\section{ADNI PET analysis}
\label{SEC:realdata}

The  dataset used in the preparation of this article were obtained from the ADNI database (\url{adni.loni.usc.edu}).
The ADNI is a longitudinal multicenter study designed to develop clinical, imaging, genetic, and biochemical biomarkers for the early detection and tracking of AD.
From this database,  we collect PET data from $79$ patients in AD group. This  PET  dataset has been  spatially normalized and post-processed. These AD patients have three to six times doctor visits and we only select the PET scans obtained in the third visits. Patients' age ranges from $59$ to $88$ and average age is $76.49$. There are $33$ females and $46$ males among these $79$ subjects.  All scans were reoriented into $79\times 95 \times 69$ voxels, which means each patient has $69$ sliced 2D images with  $79\times 95$ pixels.  For 2D case, it means each subject has $N=7,505=79\times 95$ observed pixels for each selected image slice.   For 3D case, the observed number of voxels for each patient's brain sample is $N=79\times 95 \times 69$, which is more than $0.5$ million.

 \subsection{2D case}
\begin{figure}
\begin{center}
\hspace{0.5cm}\textbf{$20$-th} \hspace{3cm}\textbf{$40$-th}\hspace{3cm}\textbf{$60$-th}\\
Avg
$\begin{array}{l}
\includegraphics[width = .25\textwidth]{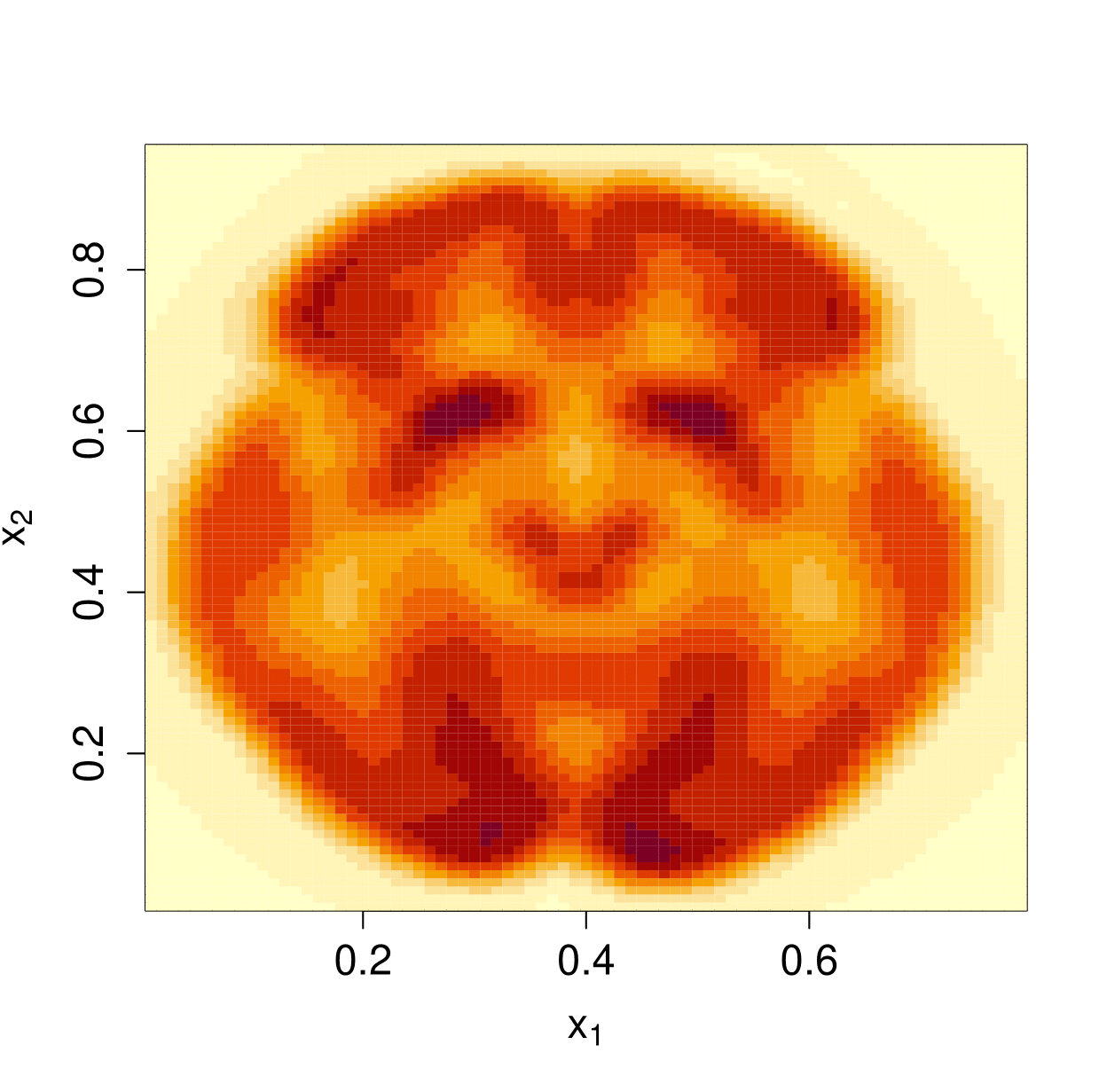}  
\hspace{2mm}
\includegraphics[width = .25\textwidth]{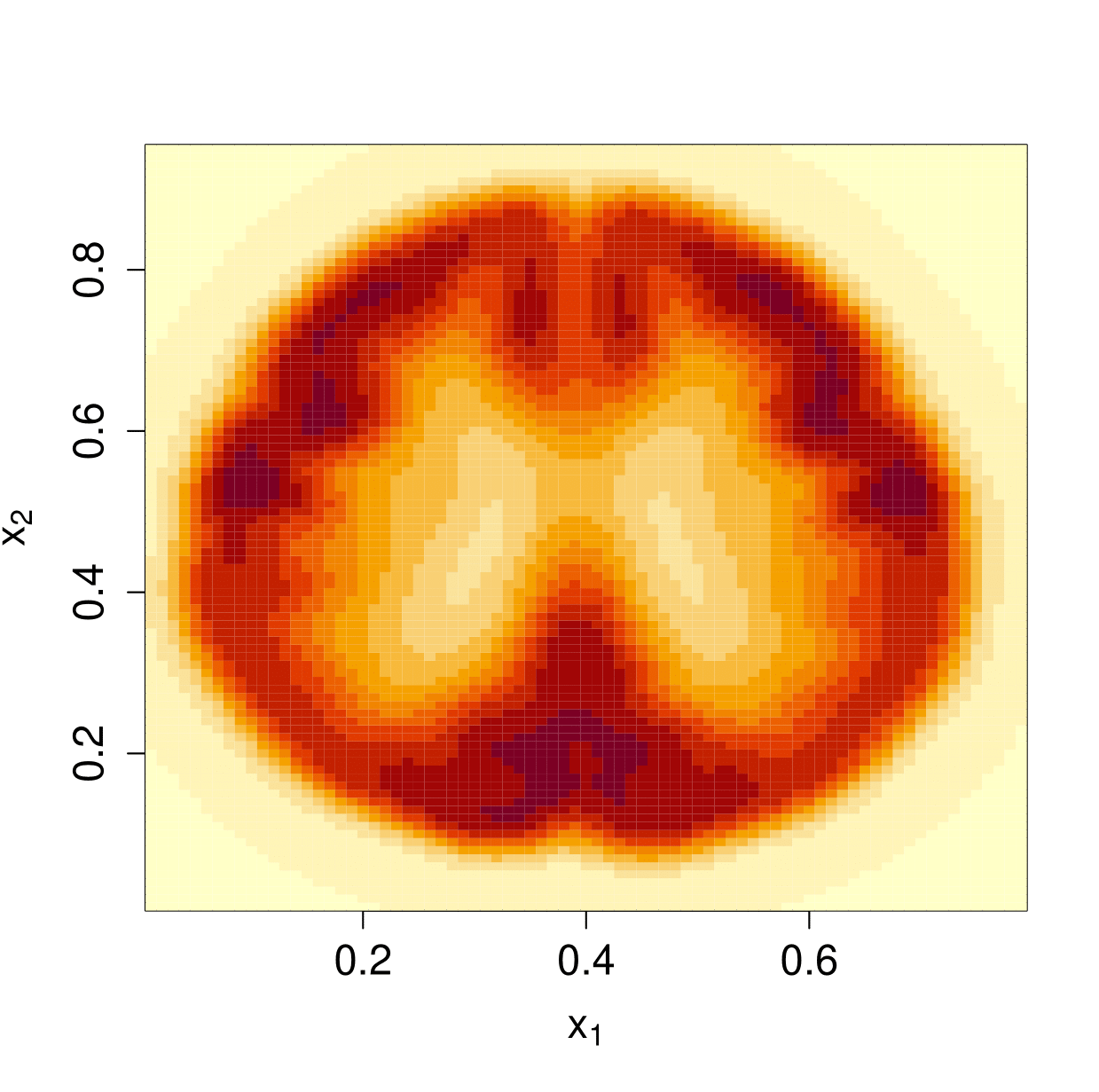}
\hspace{2mm}
\includegraphics[width = .25\textwidth]{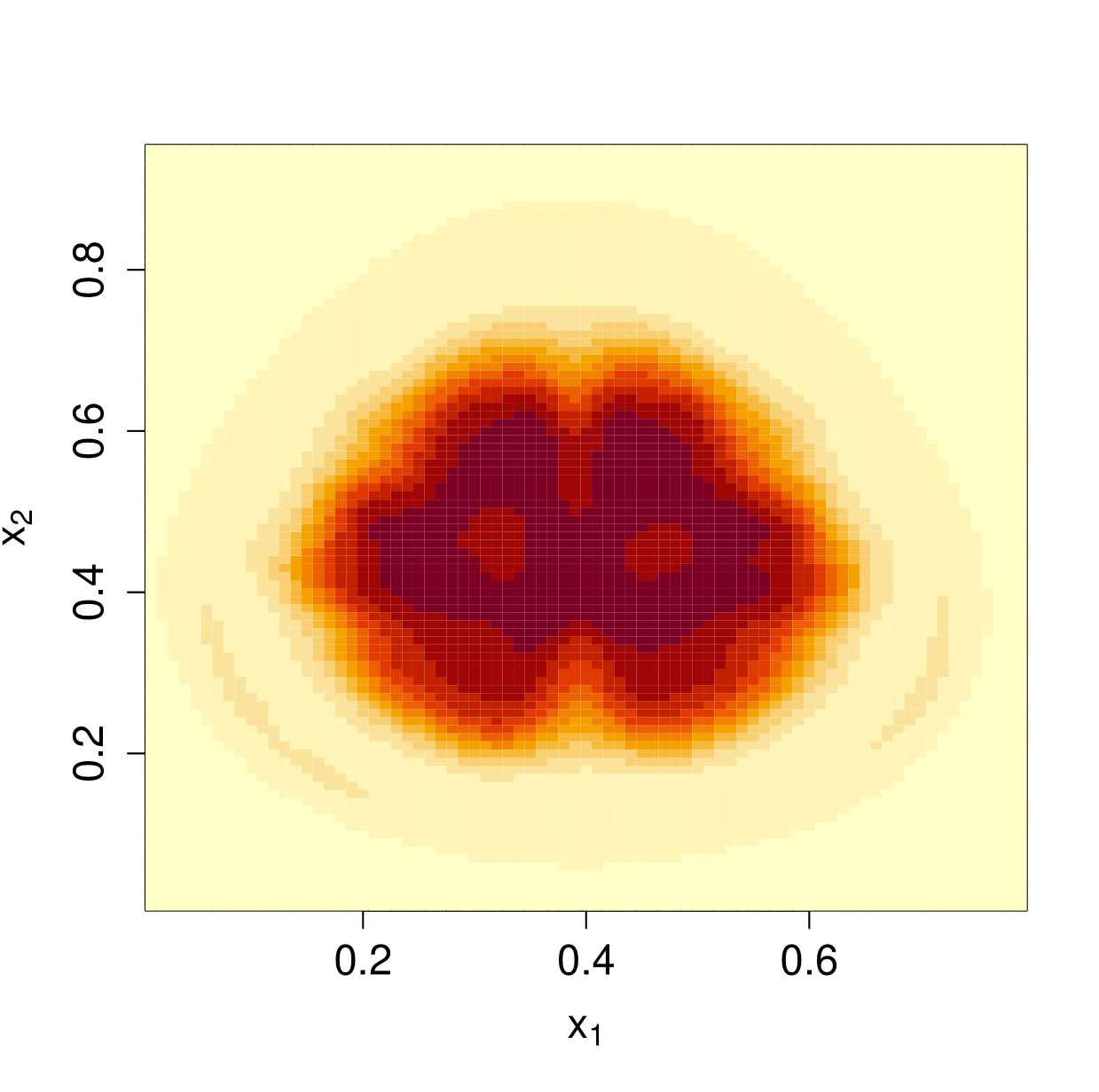} \\
\end{array}$\\
Low
$\begin{array}{l}
\includegraphics[width = .25\textwidth]{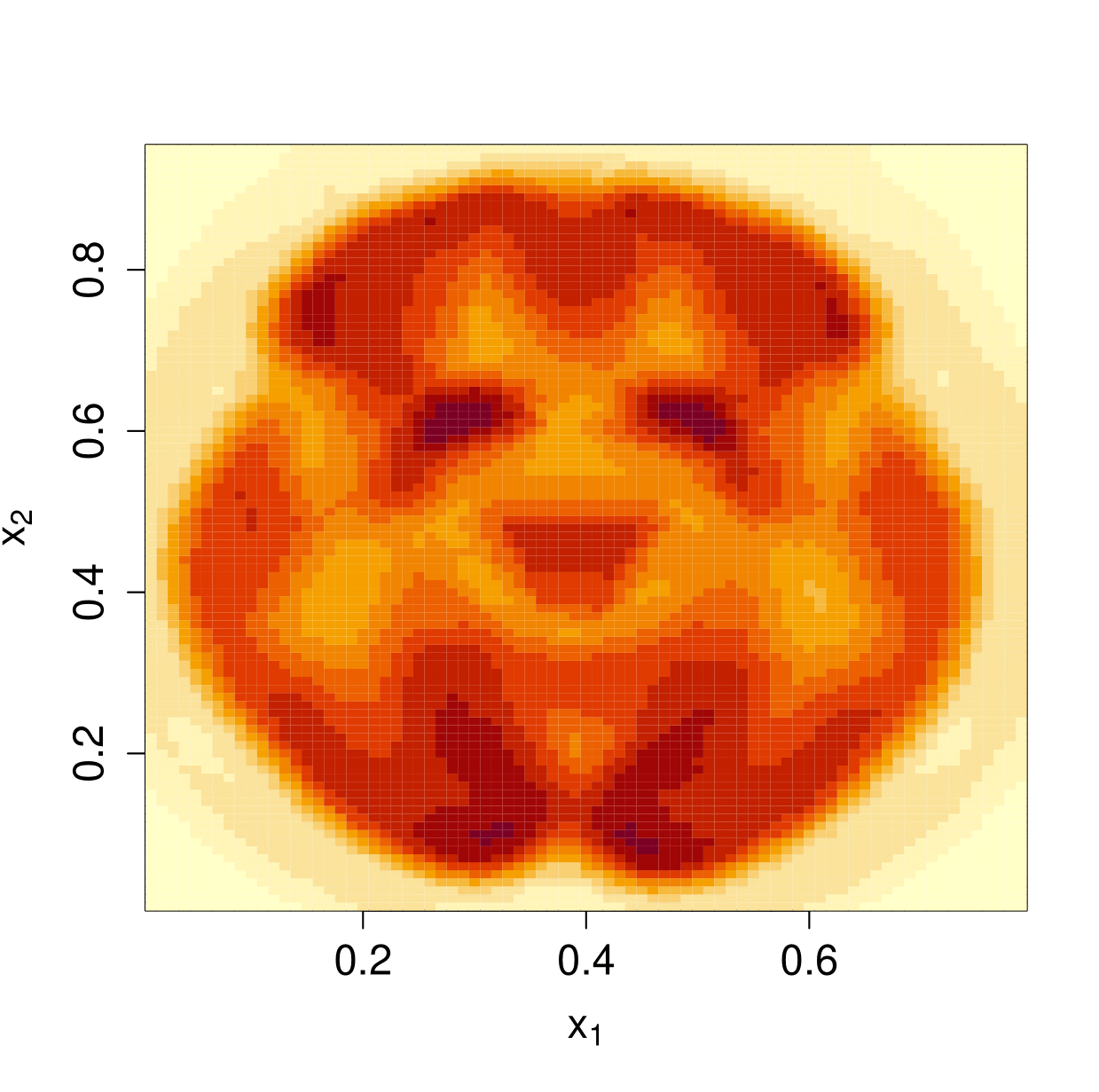} 
\hspace{2mm}
\includegraphics[width = .25\textwidth]{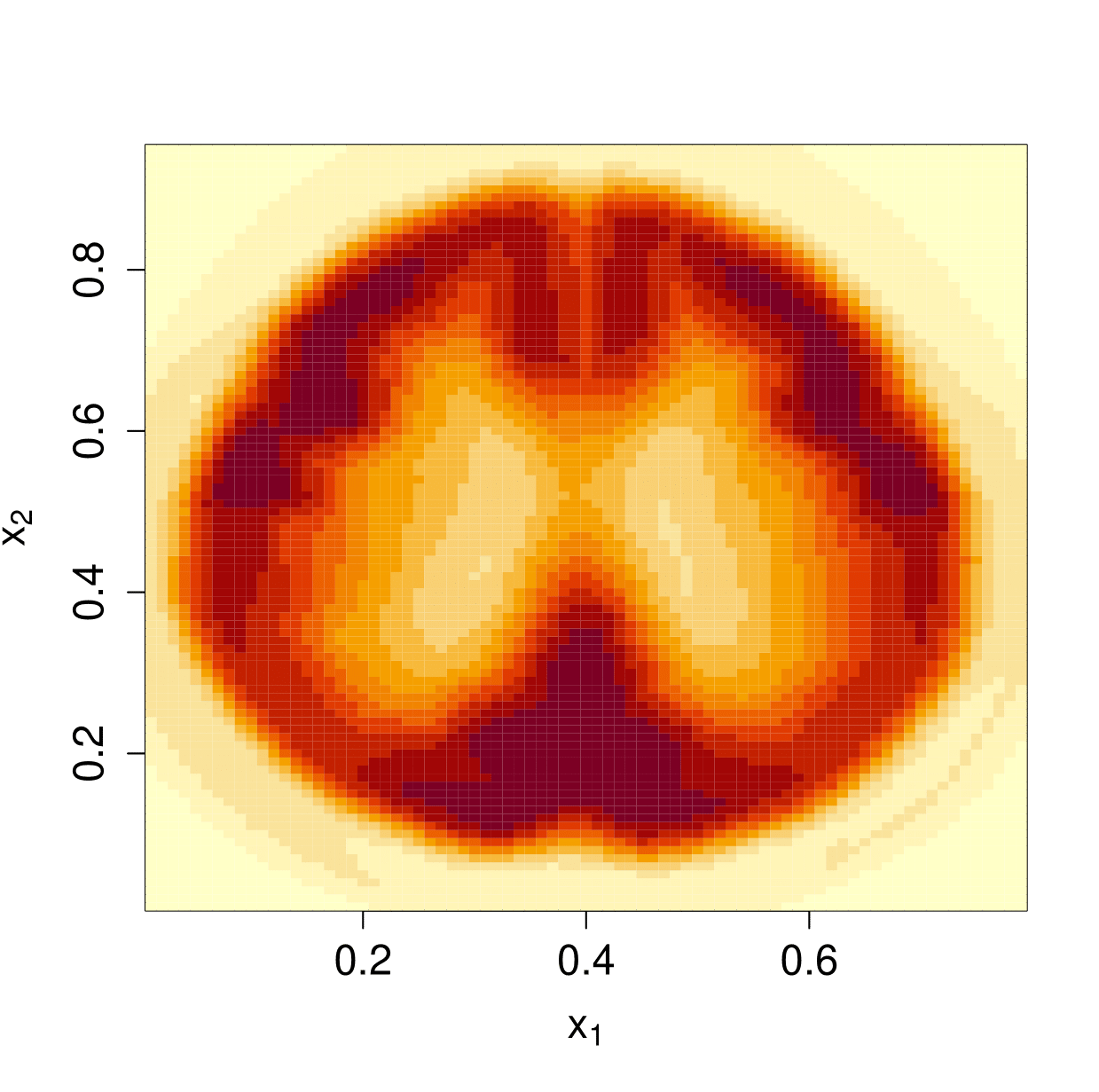}
\hspace{2mm}
\includegraphics[width = .25\textwidth]{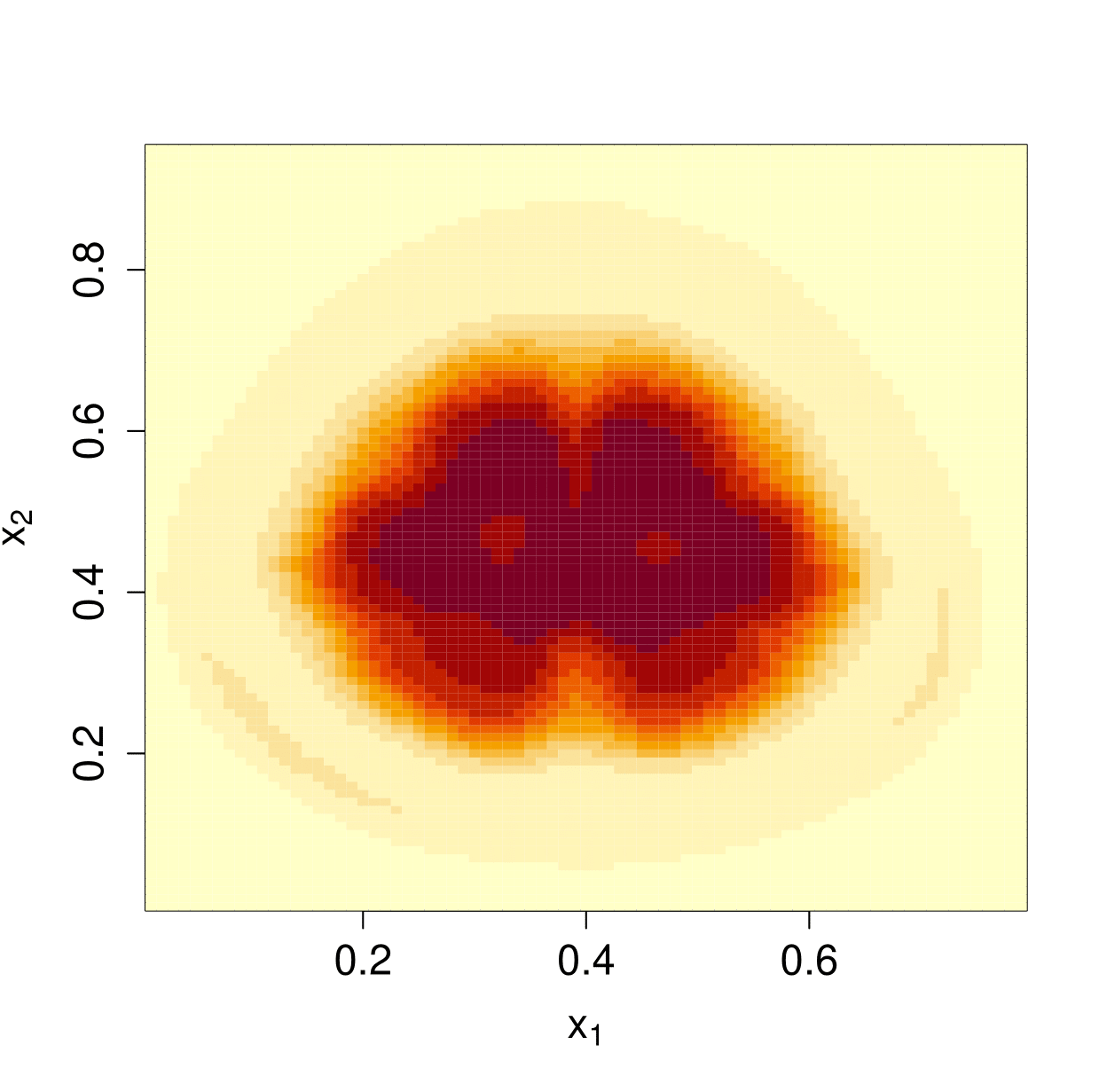} \\
\end{array}$\\
High
$\begin{array}{l}
\includegraphics[width = .25\textwidth]{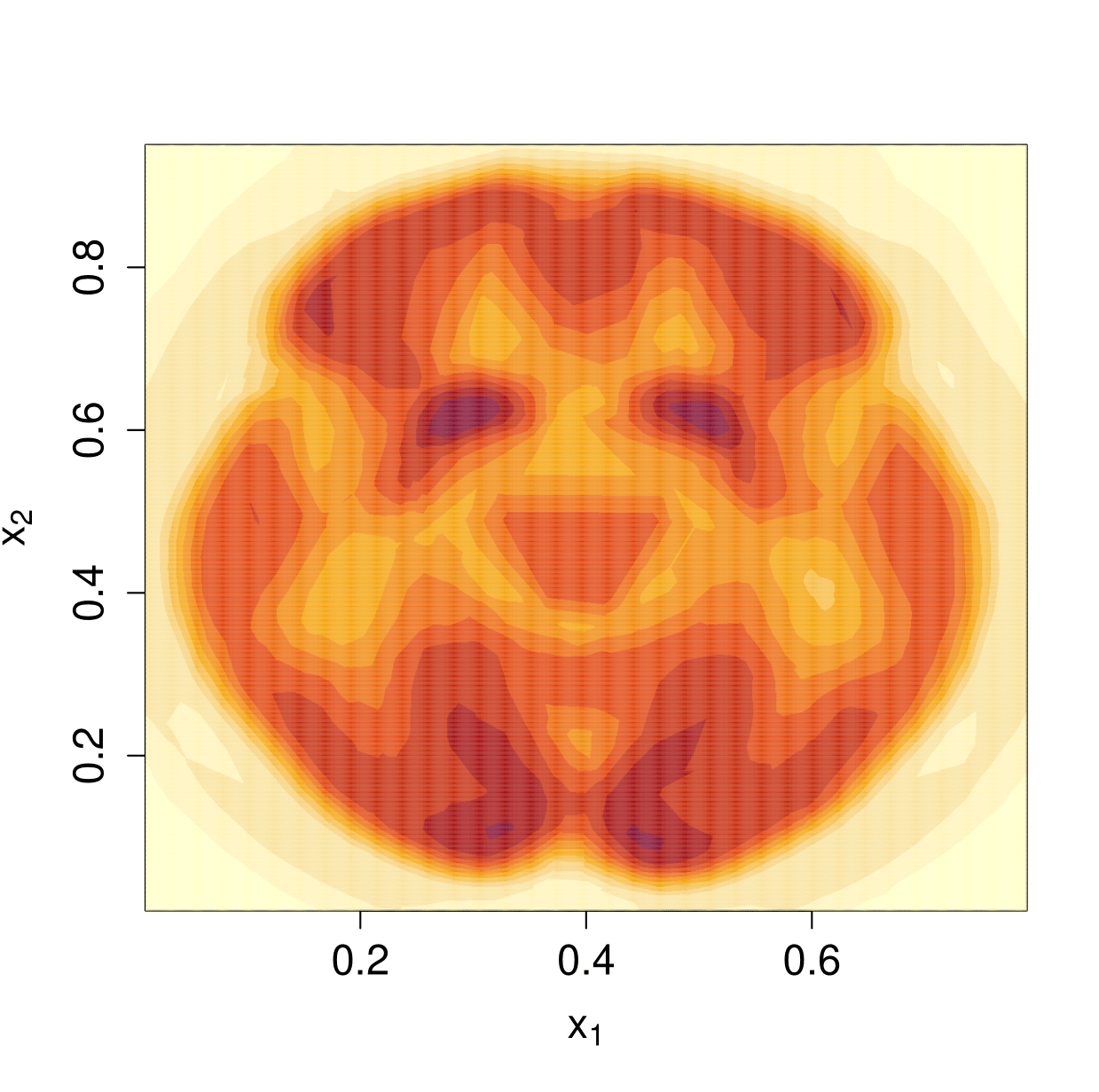}
\hspace{1.mm}
\includegraphics[width = .25\textwidth]{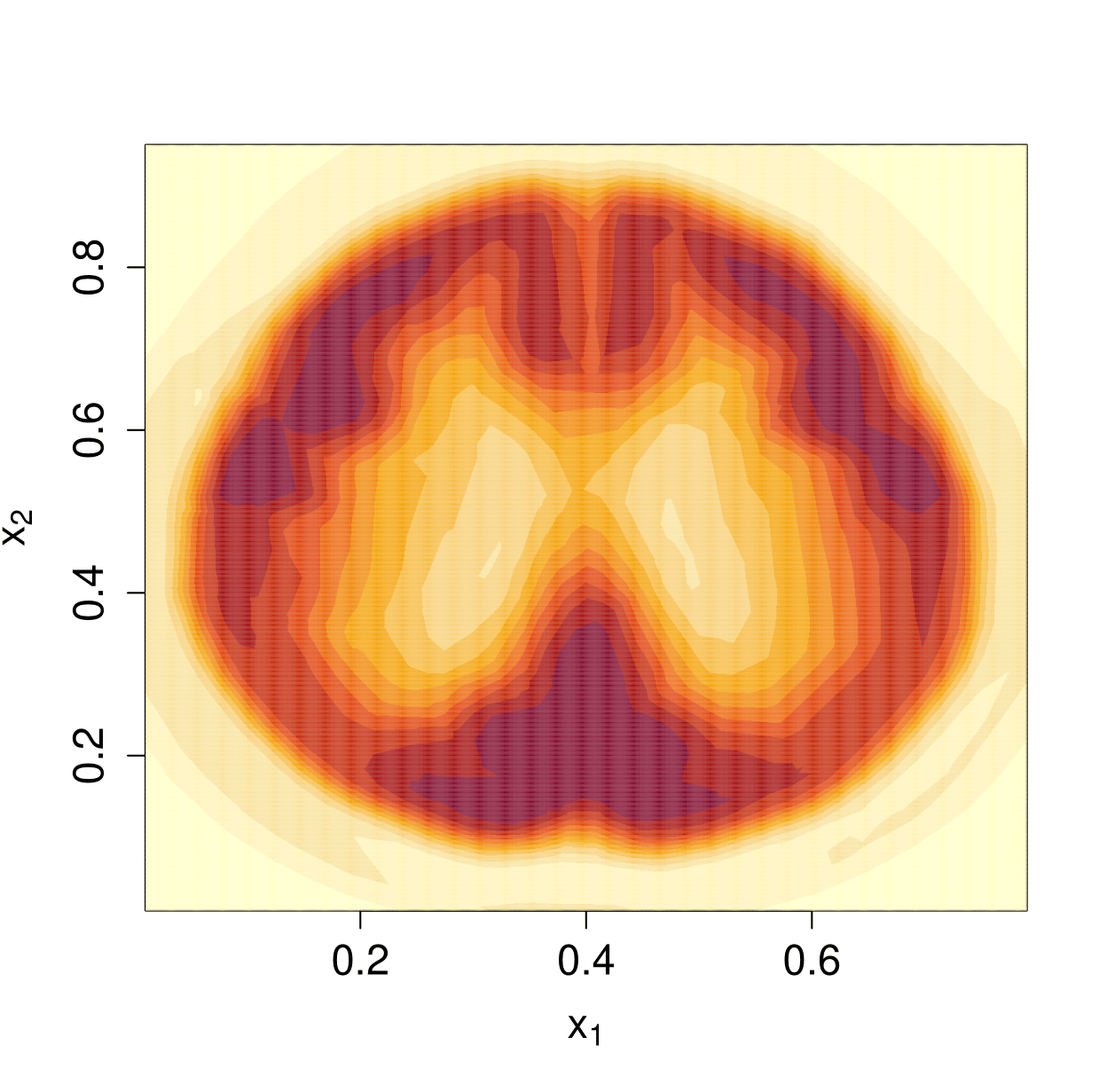}
\hspace{1.mm}
\includegraphics[width = .25\textwidth]{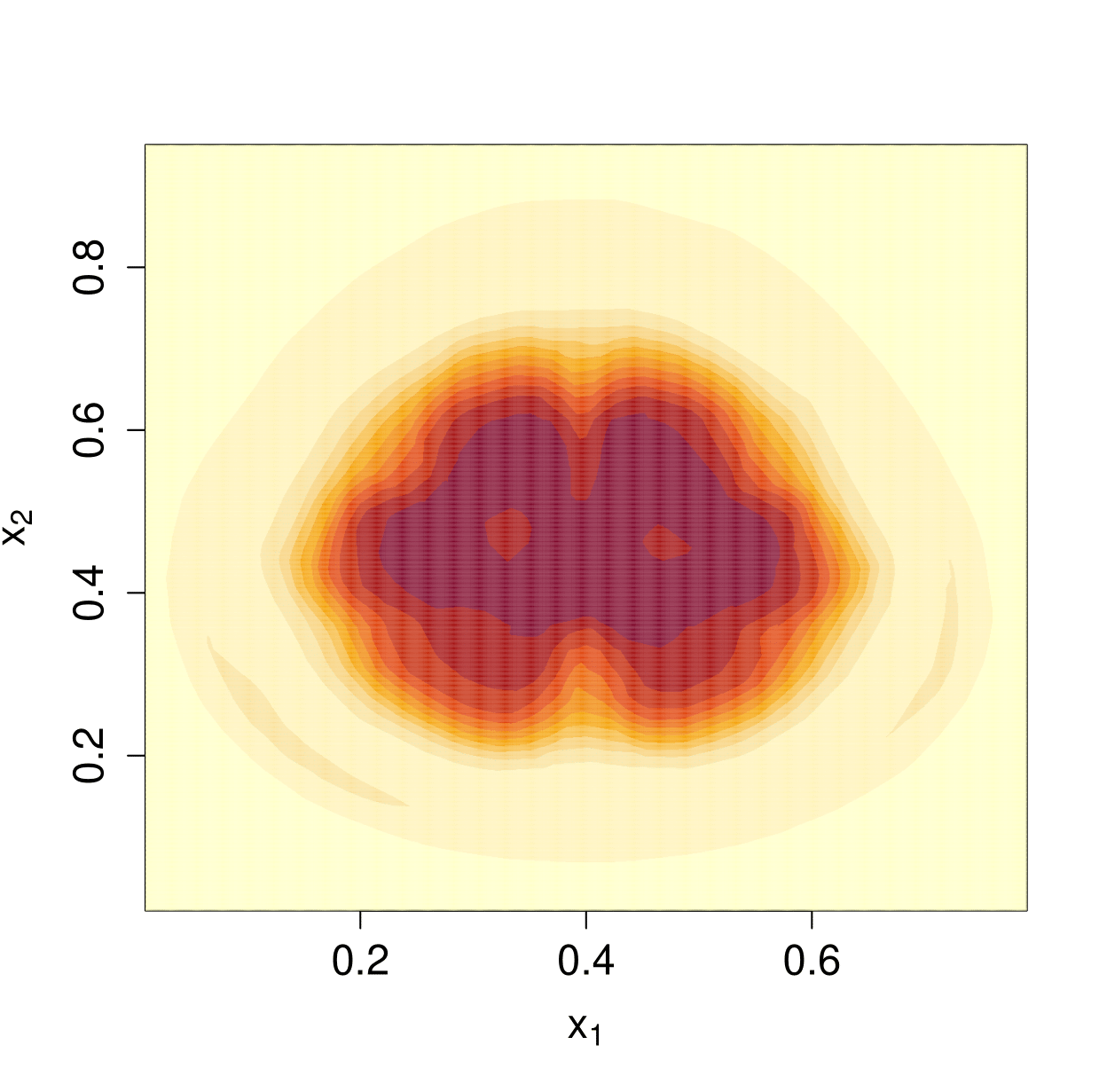} \\ 
\end{array}$\\
\caption{ From top to bottom are averaged images $\{\overline{Y}_{\cdot j}\}_{j=1}^{7505}$, recovered images $\widehat{f}(x_{1j},x_{2j'})$, $j=1,\ldots, 79$, $j'=1,\ldots, 95$ and recovered high resolution ($128\times 128$) images $\widehat{f}(x_{1j},x_{2j})$, $j=1,\ldots, 128$.   Left: The $20$-th slices; Middle: The $40$-th slices; Right: The $60$-th slices. }
\label{FIG:Realdata 2D}
\end{center}
\end{figure}

For 2D case, we select the $20$-th, $40$-th and $60$-th slices from $69$ slices for each patient. We first take average across $79$ patients for each slices (see the first row in Figure \ref{FIG:Realdata 2D}).  Then, based on the averaged images, we obtain the proposed DNN estimators for each  slice (see the second row in Figure \ref{FIG:Realdata 2D}). We also recover the image with higher resolutions $512\times 512$ pixels, instead of the original  $95 \times 69$ pixels for each slice (see the third row in Figure \ref{FIG:Realdata 2D}).

\subsection{3D case}
\begin{figure}
\begin{center}
\hspace{0.3cm}\textbf{$20$-th} \hspace{3.5cm}\textbf{$40$-th}\hspace{3.5cm}\textbf{$60$-th}\\
\includegraphics[width = .30\textwidth]{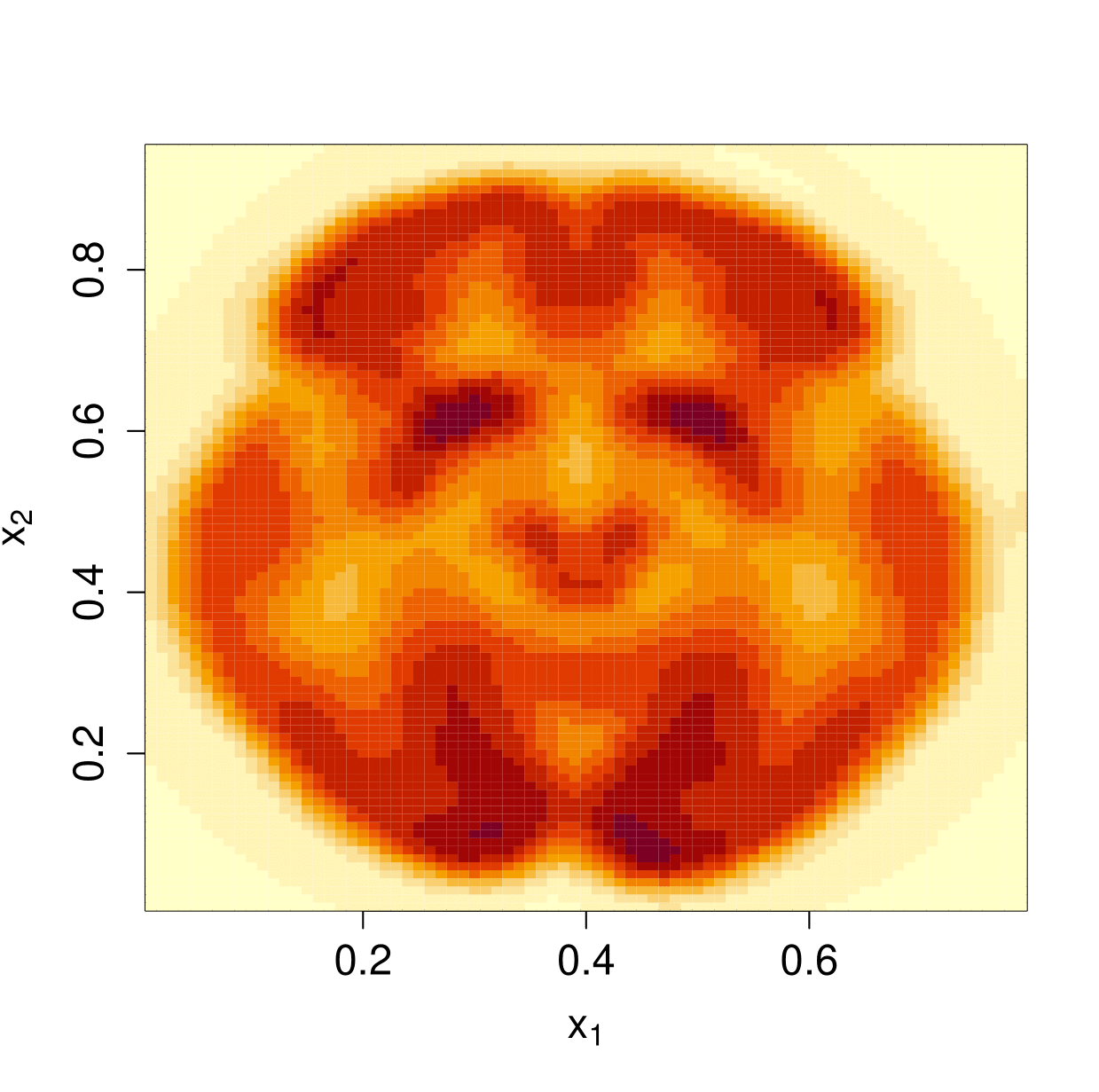}  
\includegraphics[width = .30\textwidth]{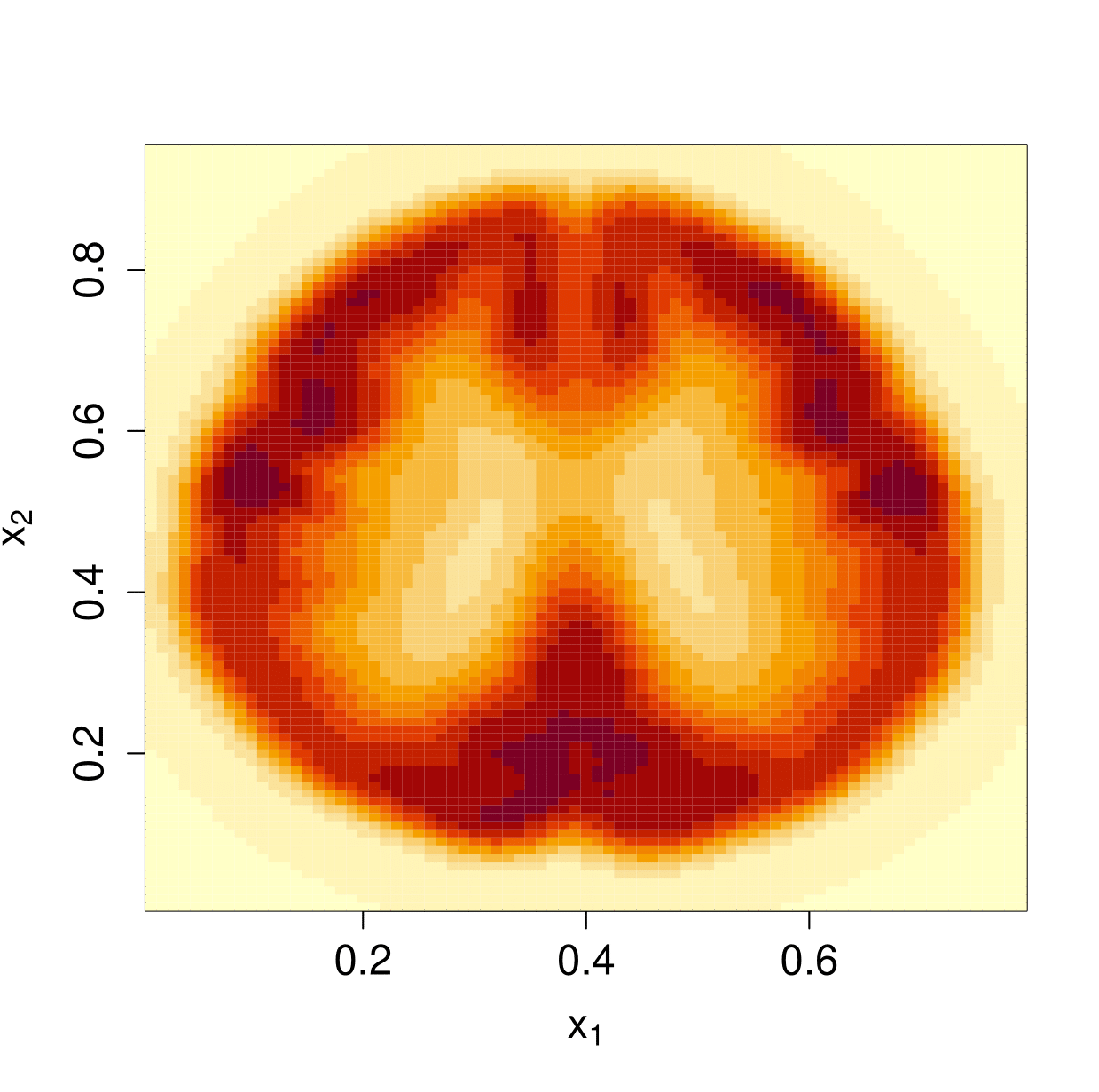}
\includegraphics[width = .30\textwidth]{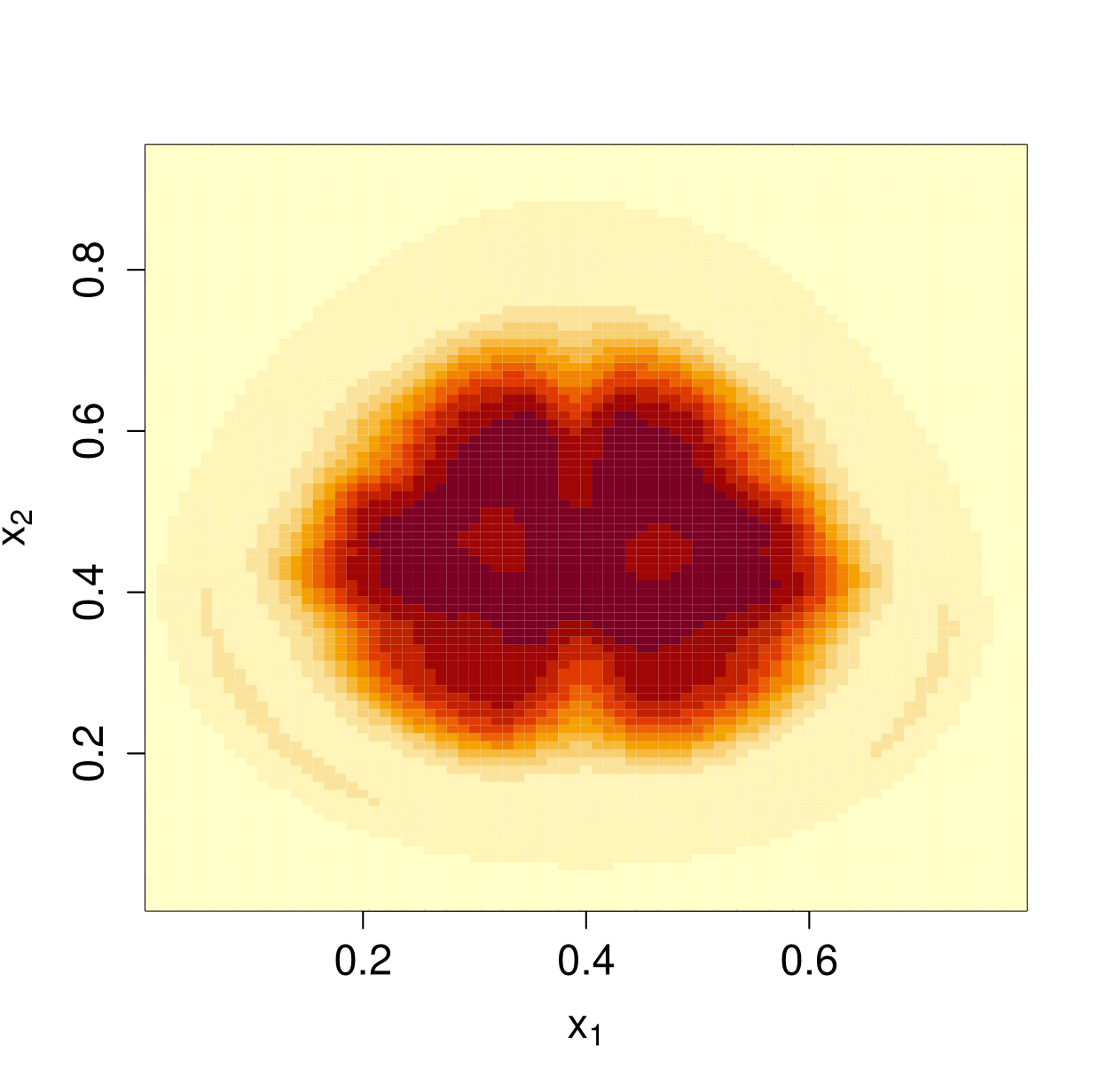}
\caption{Recovered $20$-th, $40$-th and $60$-th slices in 3D case.}
\label{FIG:Realdata 3D recover}
\end{center}
\end{figure}

\begin{figure}
\begin{center}
\textbf{$1$-st} \hspace{3.6cm}\textbf{$17$-th}\hspace{3.6cm}\textbf{$33$-th}\\
\includegraphics[width = .30\textwidth]{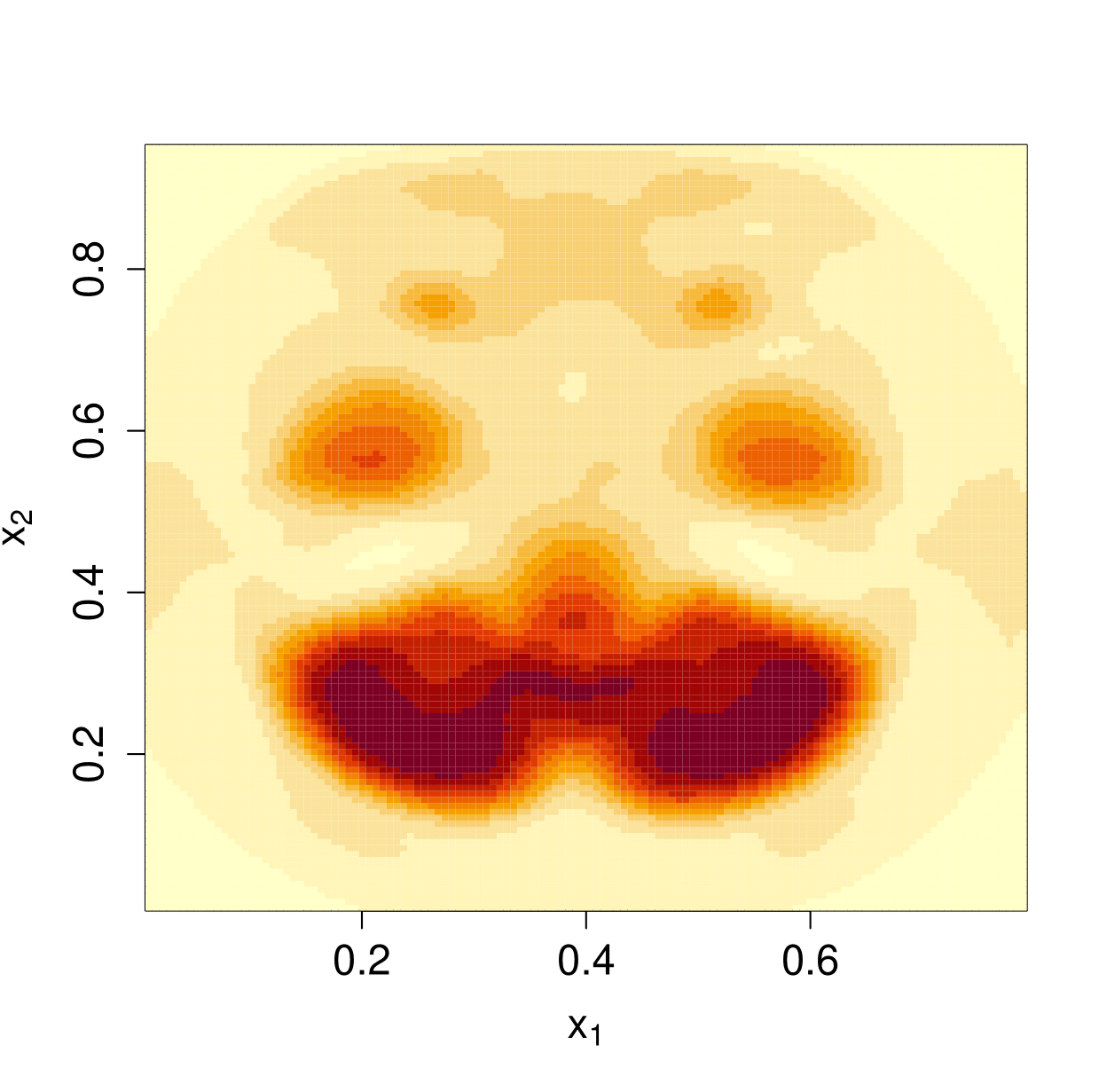}  
\includegraphics[width = .30\textwidth]{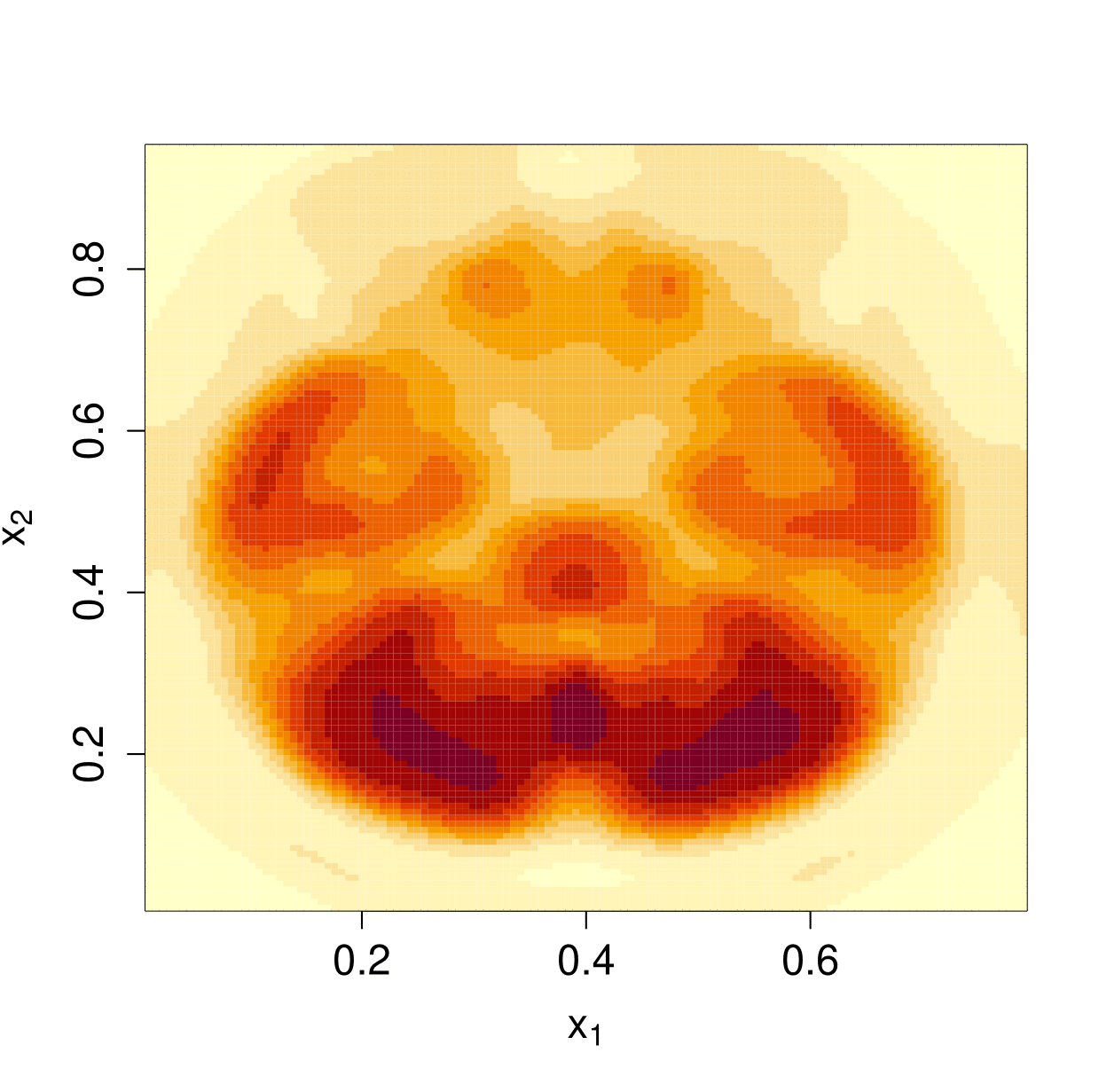}
\includegraphics[width = .30\textwidth]{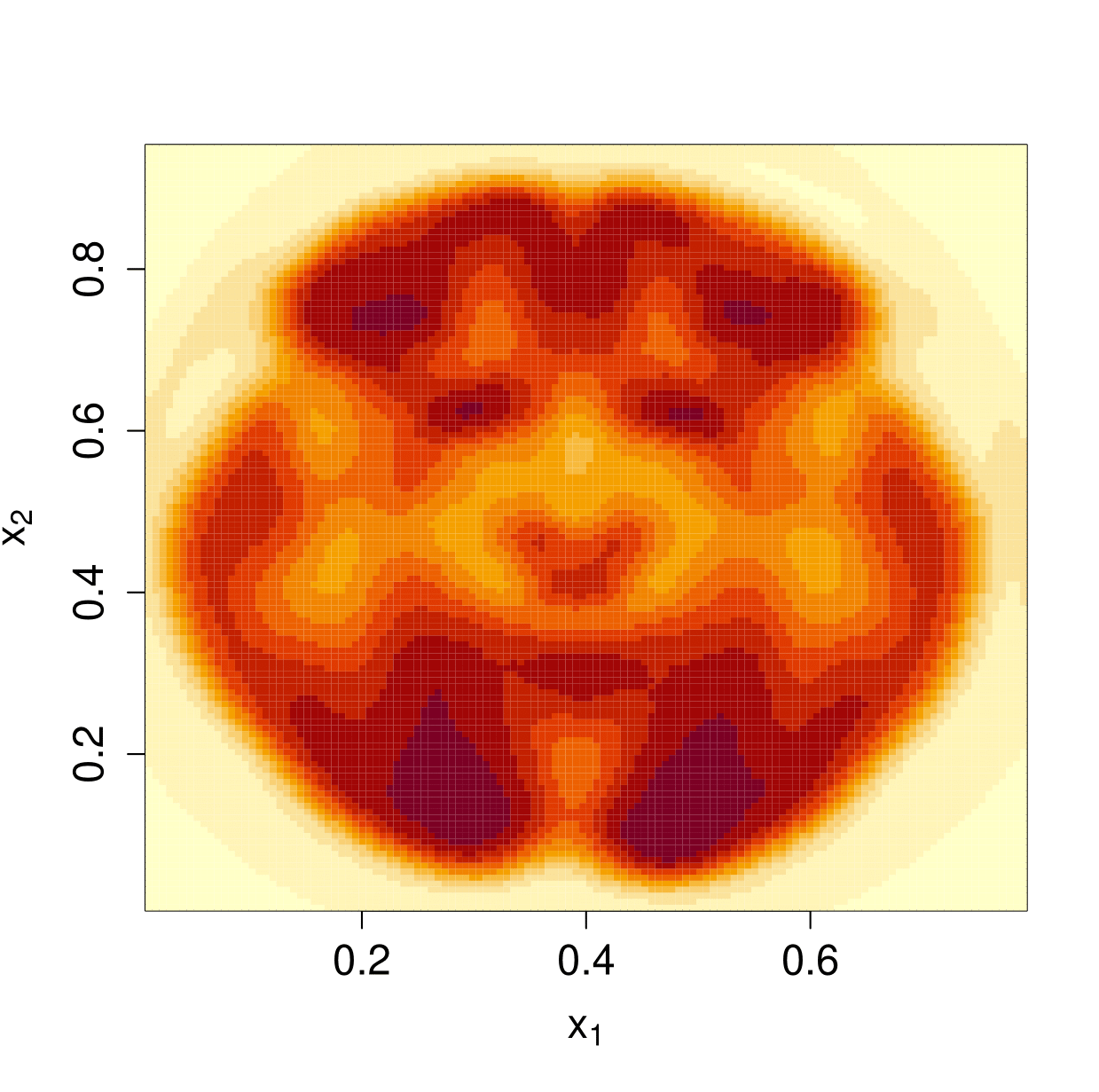}\\
\textbf{$49$-th} \hspace{3.6cm}\textbf{$65$-th}\hspace{3.6cm}\textbf{$81$-th}\\
\includegraphics[width = .30\textwidth]{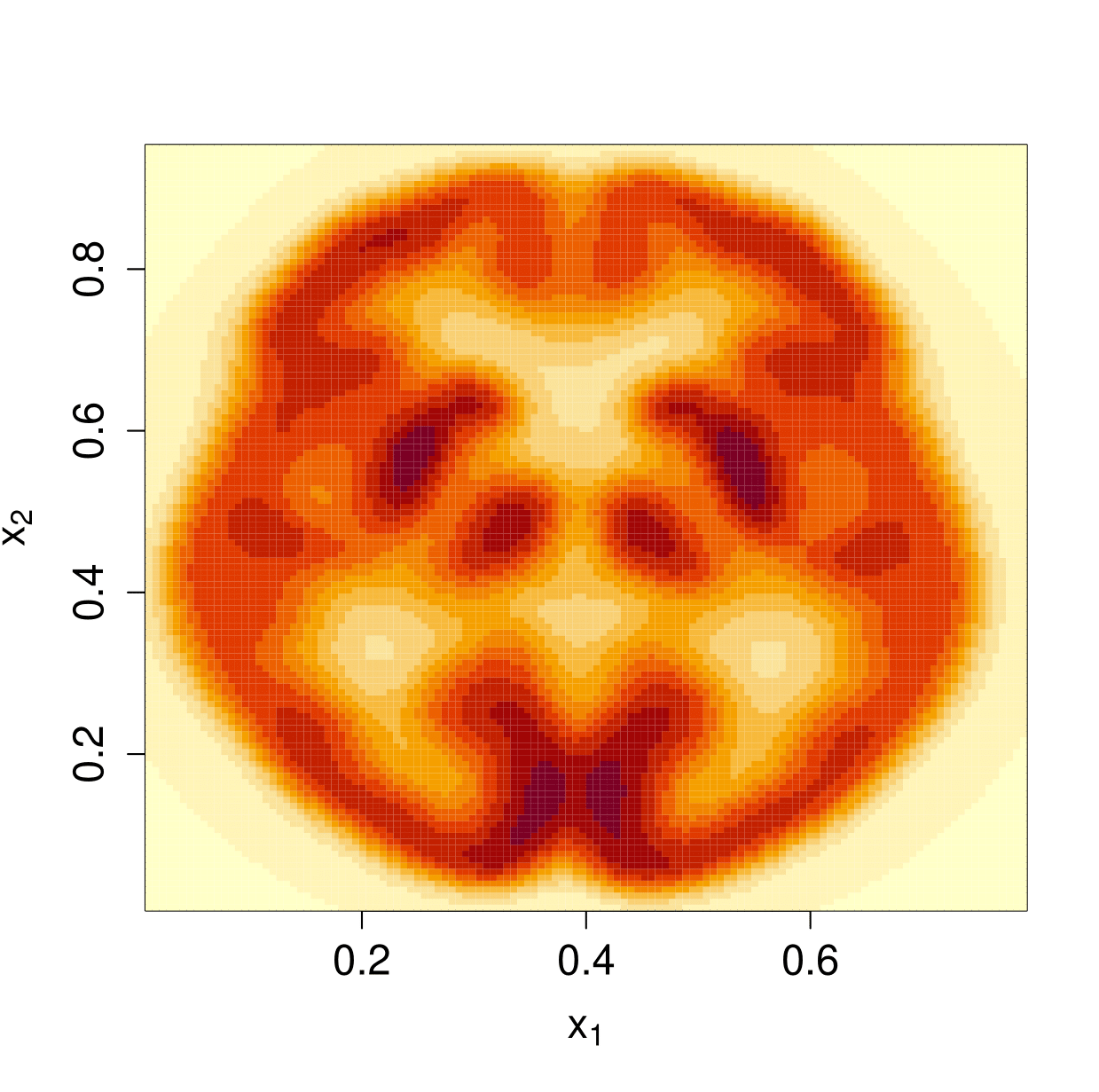}  
\includegraphics[width = .30\textwidth]{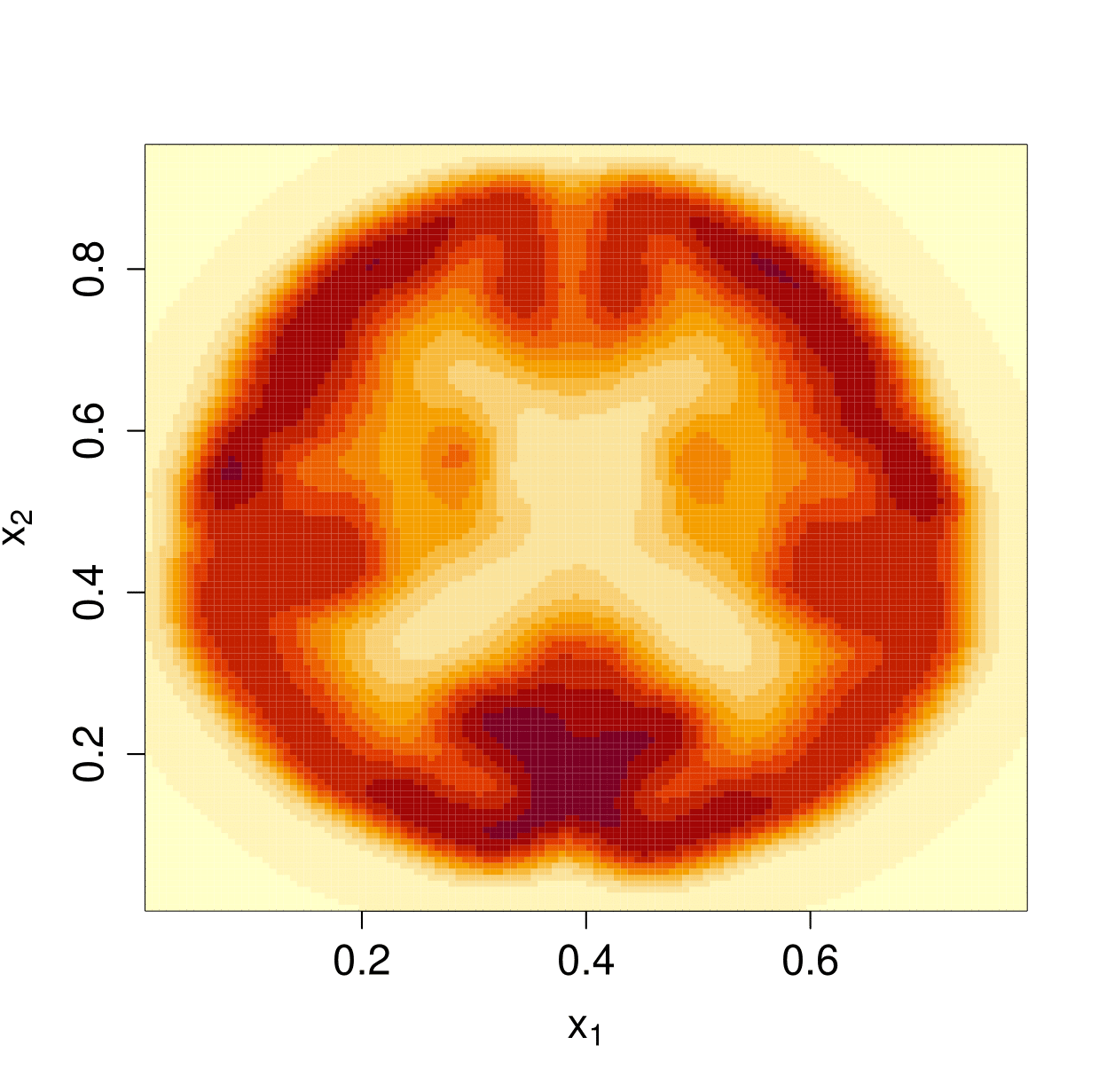}
\includegraphics[width = .30\textwidth]{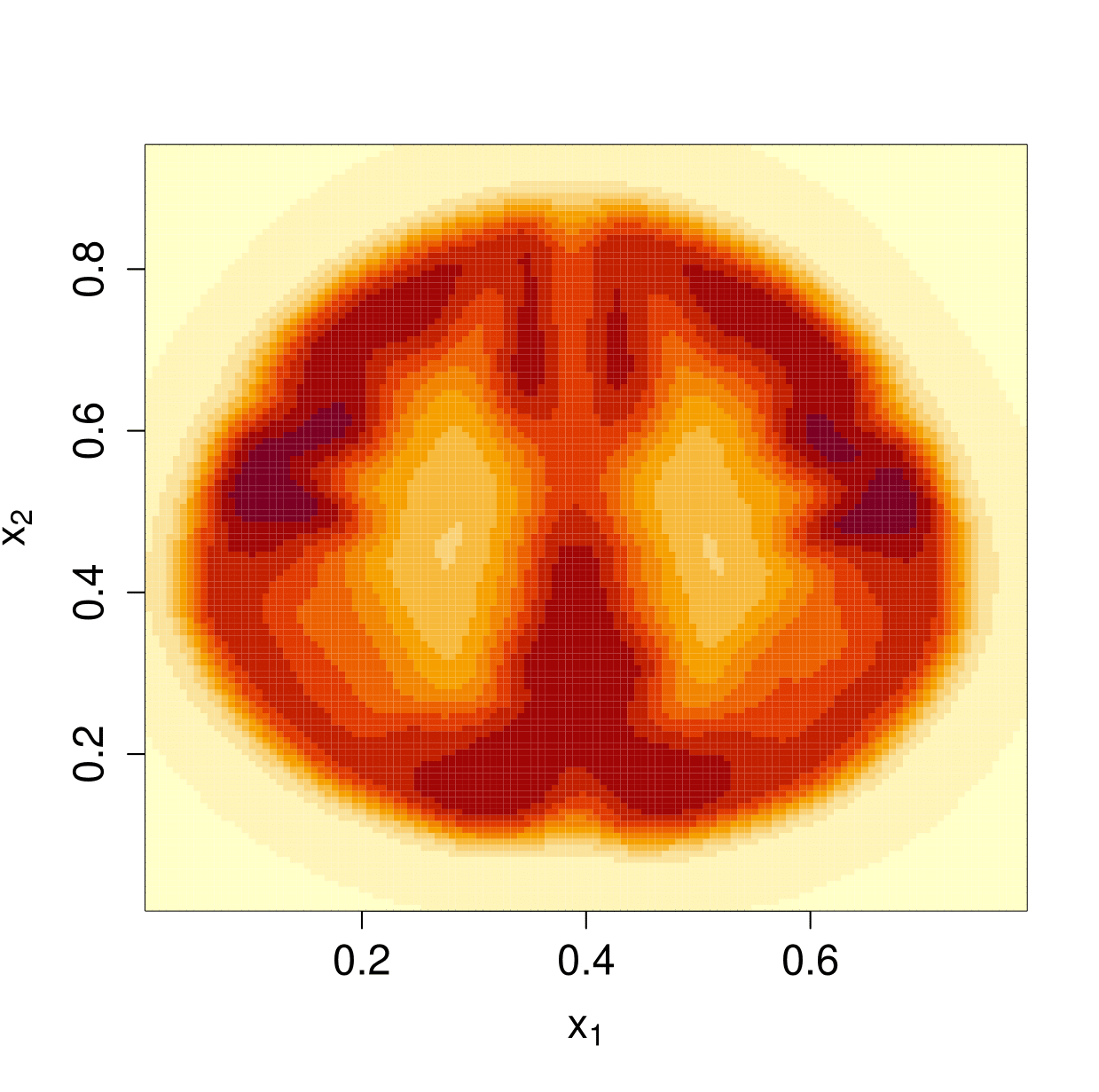}\\
\textbf{$97$-th} \hspace{3.5cm}\textbf{$113$-th}\hspace{3.5cm}\textbf{$128$-th}\\
\includegraphics[width = .30\textwidth]{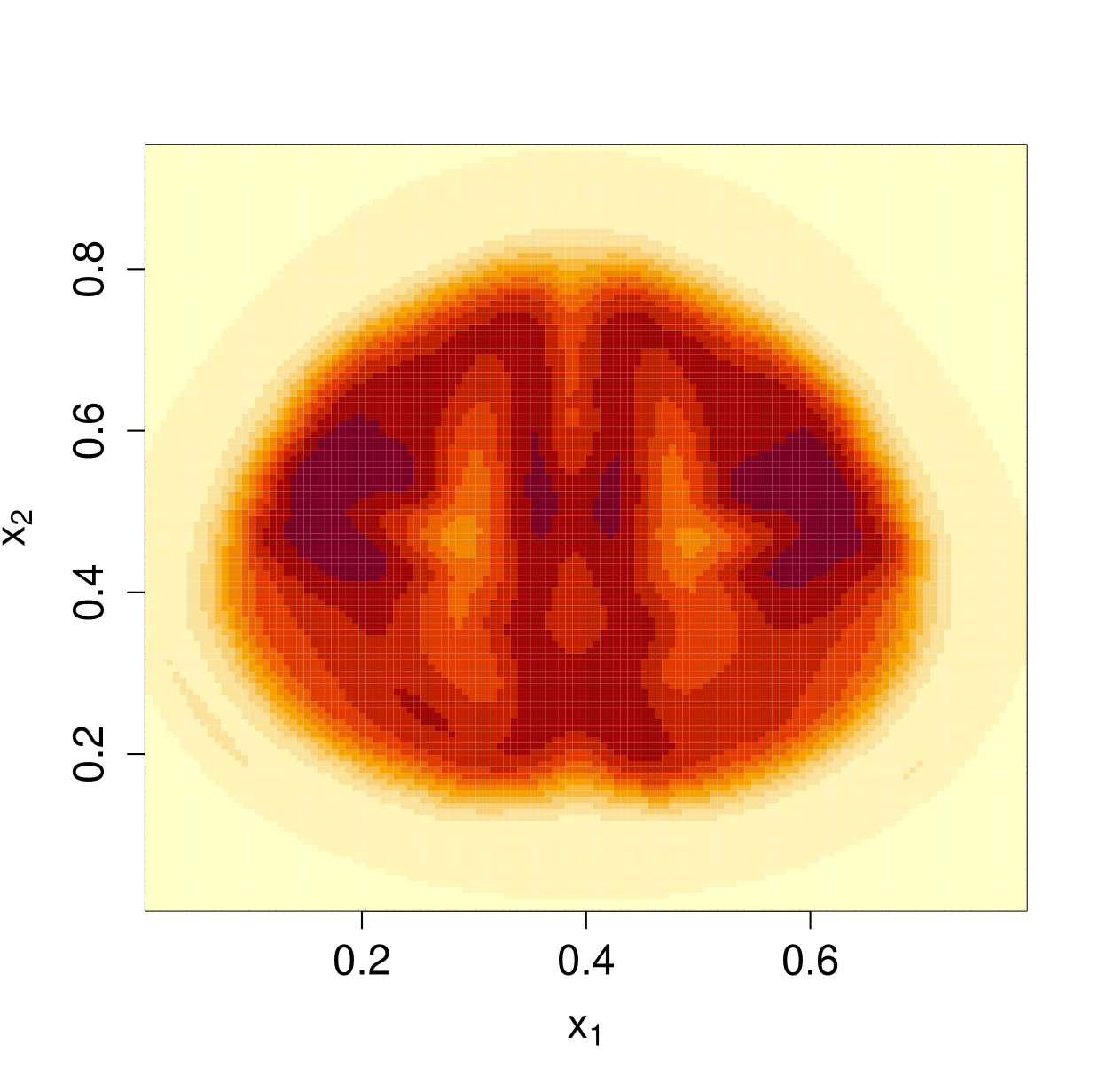}  
\includegraphics[width = .30\textwidth]{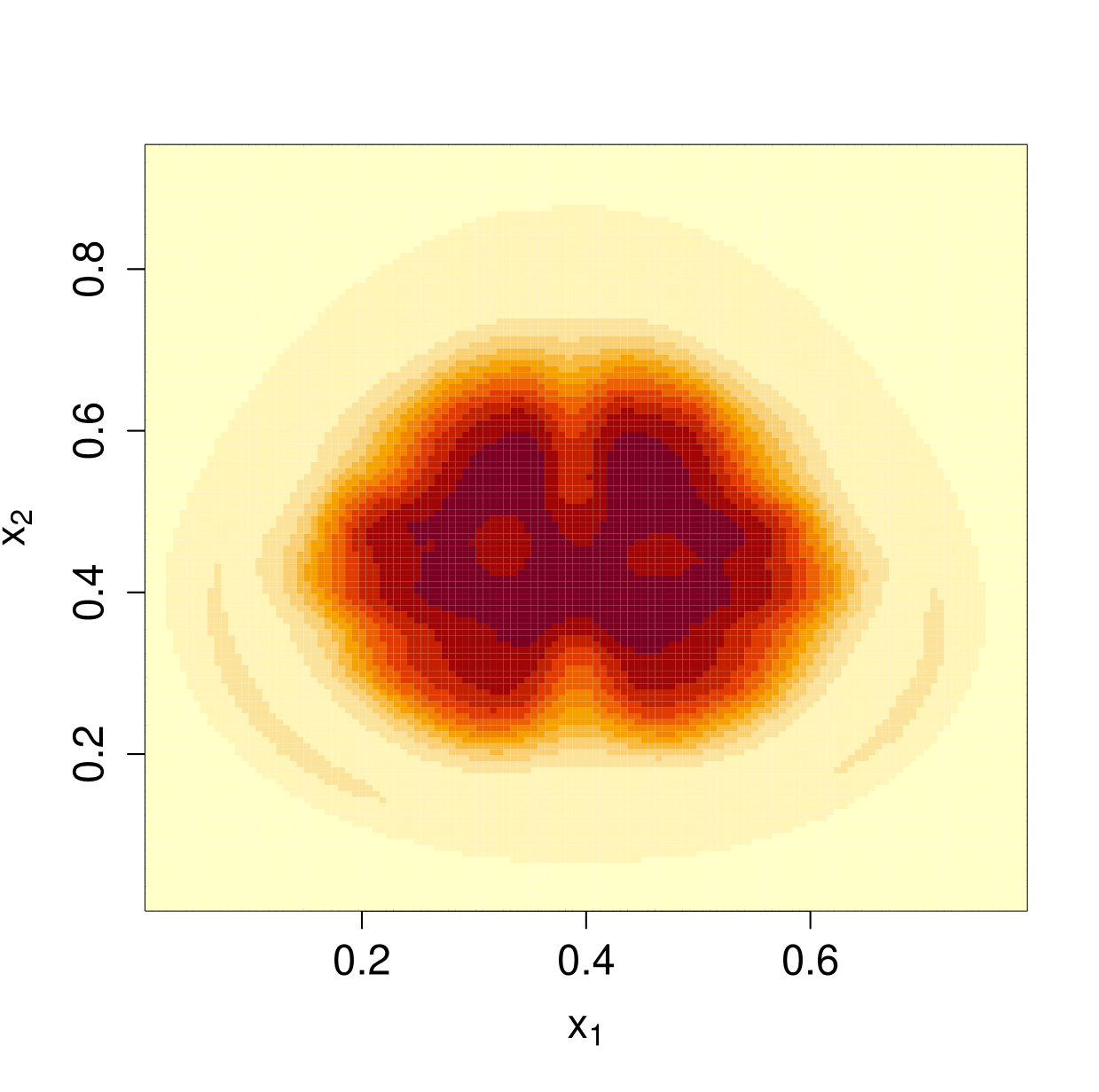}
\includegraphics[width = .30\textwidth]{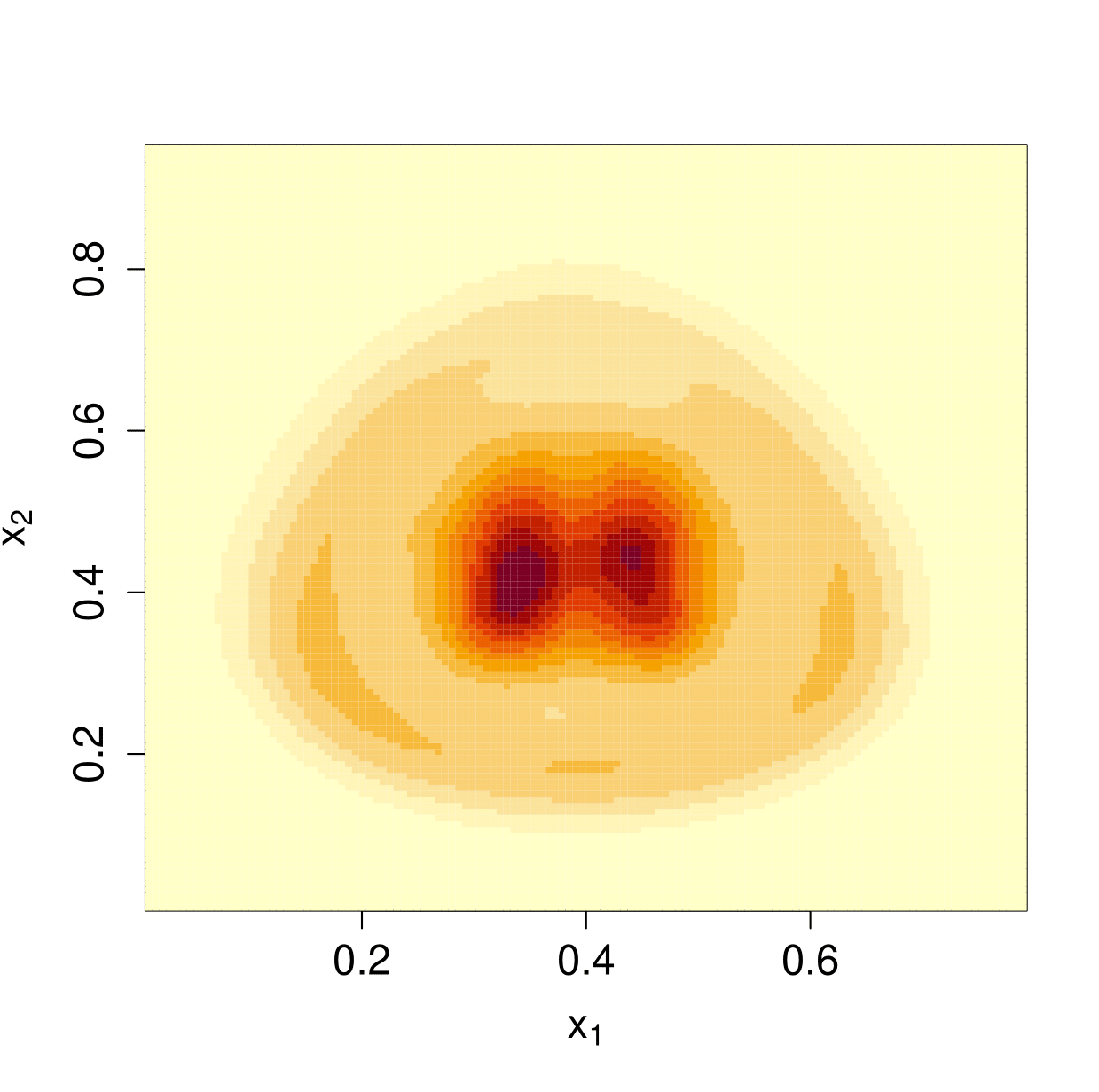}
\caption{Recovered higher resolutions of selected nine slices in 3D case.}
\label{FIG:Realdata 3D recover high}
\end{center}
\end{figure}

In 3D case, on $79$ patients, and total $79\times 95 \times 69$ voxels. Same as 2D case, we first average the total $79$ 3D scans into one 3D scan, and then perform neural network to train the model based on the averaged 3D image. In Figure \ref{FIG:Realdata 3D recover}, we break down the recovered 3D image and show the recovered $20$-th, $40$-th and $60$-th slices.
In Figure \ref{FIG:Realdata 3D recover high}, we also recover the image in higher resolutions $128 \times 128 \times 128$ voxels, which means instead of the original $79\times 95 \times 69$ voxels, we can provide the estimated image slices with higher resolution ($128 \times 128$ pixels, instead of the original $79 \times 95$ pixels) at finer grid points ($128$ points, instead of the original $69$ points).

\section{Discussion}\label{SEC: discussion}
In this work, we resolve the curse-of-dimensionality and model misspecification issues in high-dimensional FDA via the promising technique from the deep learning domain.
 By properly choosing network
architecture, our estimator achieves the optimal nonparametric convergence rate in empirical norm. Under certain circumstances such as
trigonometric polynomial kernel and a sufficiently large sampling frequency, the convergence rate of the proposed DNN estimator
is even faster than root-$n$ rate. To our best knowledge, this is the first piece of work in FDA, which yields attractive empirical convergence rate for high-dimensional FDA, and at meanwhile is free from curse-of-dimensionality and model misspecification.  Numerical analysis demonstrates that our
approach is useful in recovering the signal for high-dimensional imaging data.  Some interesting future works may include the functional linear regression model and classification problems in the framework of DNN.

\section*{Acknowledgements}
Wang's and  Cao's  research was partially supported by  NSF award DMS 1736470.
Shang's research was supported in part by NSF DMS 1764280 and 1821157. Data used in preparation of this article were obtained from the Alzheimers Disease Neuroimaging Initiative (ADNI) database (\url{adni.loni.usc.edu}). As
such, the investigators within the ADNI contributed to the design and implementation of
ADNI and/or provided data but did not participate in analysis or writing of this report.
A complete listing of ADNI investigators can be found at: \url{http://adni.loni.usc.edu/wp-content/uploads/how_to_apply/ADNI_Acknowledgement_List.pdf}.

\section{Supplementary Material}   
Supplementary material includes the proofs of  lemmas, Theorem \ref{THM: rate} and the implementation of the example 1.


\bibliographystyle{plain} 
\bibliography{Ref}       

\begin{thebibliography}{10}

\bibitem{Anthony:Bartlett:09}
M.~Anthony and P.~Bartlett.
\newblock {\em Neural Network Learning}.
\newblock Cambridge University Press, Cambridge, 2009.

\bibitem{Bauer:Kohler:19}
B.~Bauer and M.~Kohler.
\newblock On deep learning as a remedy for the curse of dimensionality in
  nonparametric regression.
\newblock {\em The Annals of Statistics}, 47:2261--2285, 2019.

\bibitem{Braun:06}
M.~L. Braun.
\newblock Accurate error bounds for the eigenvalues of the kernel matrix.
\newblock {\em Journal of Machine Learning Research}, 7:2303--2328, 2006.

\bibitem{CY:11}
T.~T. Cai and M.~Yuan.
\newblock Optimal estimation of the mean function based on discretely sampled
  functional data: Phase transition.
\newblock {\em The Annals of Statistics}, 39:2330--2355, 2011.

\bibitem{Cao:Wang:Li:Yang:14}
G.~Cao, L.~Wang, Y.~Li, and L.~Yang.
\newblock Oracle-efficient confidence envelopes for covariance functions in
  dense functional data.
\newblock {\em Statistica Sinica}, 26:359--383, 2016.

\bibitem{Cao:Yang:Todem:12}
G.~Cao, L.~Yang, and D.~Todem.
\newblock Simultaneous inference for the mean function of dense functional
  data.
\newblock {\em Journal of Nonparametric Statistics}, 24:359--377, 2012.

\bibitem{cardot:00}
Herv\'{e} Cardot.
\newblock Nonparametric estimation of smoothed principal components analysis of
  sampled noisy functions.
\newblock {\em Journal of Nonparametric Statistics}, 12:503--538, 2000.

\bibitem{Chen:Jiang:17}
Lu-Hung Chen and Ci-Ren Jiang.
\newblock Multi-dimensional functional principal component analysis.
\newblock {\em Stat. Comput.}, 27(5):1181--1192, 2017.

\bibitem{ferraty2006nonparametric}
Fr\'ed\'eric Ferraty and Philippe Vieu.
\newblock {\em Nonparametric functional data analysis: Theory and practice}.
\newblock Springer Series in Statistics. Springer, New York, 2006.

\bibitem{Hall:etal:06}
P.~Hall, H.~G. M\"{u}ller, and J.~L. Wang.
\newblock Properties of principal component methods for functional and
  longitudinal data analysis.
\newblock {\em The Annals of Statistics}, 34:1493--1517, 2006.

\bibitem{Kingma:Ba:15}
D.~Kingma and J.~Ba.
\newblock Adam: A method for stochastic optimization.
\newblock {\em In the 3rd International Conference on Learning Representations
  (ICLR)}, 2015.

\bibitem{Li:Hsing:10a}
Y.~Li and T.~Hsing.
\newblock Uniform convergence rates for nonparametric regression and principal
  component analysis in functional/longitudinal data.
\newblock {\em The Annals of Statistics}, 38:3321--3351, 2010.

\bibitem{Lila:etal:16}
E.~Lila, J.~A.~D. Aston, and L.~Sangalli.
\newblock Smooth principal component analysis over two-dimensional manifolds
  with an application to neuroimaging.
\newblock {\em The Annals of Applied Statistics}, 10(4):1854--1879, 2016.

\bibitem{Liu:19}
R.~Liu, B.~Boukai, and Z.~Shang.
\newblock Optimal nonparametric inference via deep neural network.
\newblock {\em Preprint}, 2019.

\bibitem{Morris:Carroll:06}
J.~S. Morris and R.~J. Carroll.
\newblock Wavelet-based functional mixed models.
\newblock {\em Journal of the Royal Statistical Society, Series~B},
  68:179--199, 2006.

\bibitem{Ramsay:Silverman:05}
J.~O. Ramsay and B.~W. Silverman.
\newblock {\em Functional Data Analysis, Second Edition}.
\newblock Springer Series in Statistics, New York, 2005.

\bibitem{Rice:Wu:01}
J.A. Rice and C.O. Wu.
\newblock Nonparametric mixed effects models for unequally sampled noisy
  curves.
\newblock {\em Biometrics}, 57:253--259, 2001.

\bibitem{Ripley:14}
B.~D. Ripley.
\newblock {\em Pattern Recognition and Neural Networks}.
\newblock Cambridge University Press, Cambridge, 2014.

\bibitem{Schmidt:19}
J.~Schmidt-Hieber.
\newblock Nonparametric regression using deep neural networks with relu
  activation function.
\newblock {\em arXiv:1708.06633}, 2019.

\bibitem{Shang:Cheng:17}
Z.~Shang and G.~Cheng.
\newblock Computational limits of a distributed algorithm for smoothing spline.
\newblock {\em Journal of Machine Learning Research}, 18:1--37, 2017.

\bibitem{Srivastava:etal:14}
N.~Srivastava, G.~Hinton, A.~Krizhevsky, I.~Sutskever, and R.~Salakhutdinov.
\newblock Dropout: a simple way to prevent neural networks from overfitting.
\newblock {\em Journal of Machine Learning Research}, 15:1929--1958, 2014.

\bibitem{Stone:82}
Charles~J. Stone.
\newblock Optimal global rates of convergence for nonparametric regression.
\newblock {\em The Annals of Statistics}, 10(4):1040--1053, 1982.

\bibitem{Taylor:09}
J.~Taylor.
\newblock Lecture notes for stats 352: Spatial statistics.
\newblock 2009.

\bibitem{Wahba:90}
G.~Wahba.
\newblock Spline models for observational data.
\newblock {\em SIAM CBMS-NSF Regional Conference Series in Applied
  Mathematics}, 59, 1990.

\bibitem{Wang:Nan:Zhu:Koeppe:14}
B.~Wang, B.~Nan, J.~Zhu, and R.~Koeppe.
\newblock Regulzarized 3d functional regression for brain image data via haar
  wavelets.
\newblock {\em The Annals of Applied Statistics}, 8:1045--1064, 2014.

\bibitem{Wang:etal:16}
J.L. Wang, J.~M. Chiou, and H.~G. M\"{u}ller.
\newblock Functional data analysis.
\newblock {\em Annual Review of Statistics and Its Application}, 3:257--295,
  2016.

\bibitem{Wang:Wang:Wang:Ogden:19}
Y.~Wang, G.~Wang, L.~Wang, and T.~Ogden.
\newblock Simultaneous confidence corridors for mean functions in functional
  data analysis of imaging data.
\newblock {\em Biometrics}, page In press, 2019.

\bibitem{Yao:etal:05b}
F.~Yao, H.~G. M\"{u}ller, and J.~L. Wang.
\newblock Functional data analysis for sparse longitudinal data.
\newblock {\em Journal of the American Statistical Association}, 100:577--590,
  2005.

\bibitem{Yao:etal:05a}
F.~Yao, H.~G. M\"{u}ller, and J.~L. Wang.
\newblock Functional linear regression analysis for longitudinal data.
\newblock {\em The Annals of Statistics}, 33:2873--2903, 2005.

\bibitem{Zhou:Pan:14}
Lan Zhou and Huijun Pan.
\newblock Principal component analysis of two-dimensional functional data.
\newblock {\em Journal of Computational and Graphical Statistics},
  23(3):779--801, 2014.

\end{thebibliography}

\end{document}